\pdfoutput=1
\documentclass[12pt]{article}
\usepackage{amsmath}
\usepackage[numbers]{natbib}
\usepackage{microtype}
\usepackage{graphicx}
\usepackage{subfig}
\usepackage{booktabs}
\usepackage[table]{xcolor} % per colori celle in tabelle
\usepackage{colortbl} % per colori celle in tabelle
\usepackage{bm}
\usepackage{pgf, tikz} % per grafici direttamente con latex
\usetikzlibrary{arrows,automata,fit}
\usetikzlibrary{shapes,snakes}
\usepackage{bbm} % for \mathbbm
\usepackage{amsfonts}
\usepackage{amsthm}
\usepackage{bm}
\usepackage{verbatim}
\newtheorem{definition}{Definition}

\usepackage{algorithm}
\usepackage{algorithmic}

\usepackage{multirow}
\newcommand{\xx}{1}
\newcommand{\yy}{1}
\newcommand{\sage}[2]{\tikz{\node[shape=circle,draw,inner sep=1pt,minimum width = 0.6cm, fill=#1]{$v_{#2}$};}}

\newcommand{\leaf}{\tikz{\node[shape=circle,draw,inner sep=1.5pt,fill=white] {};}}

\newcommand\independent{\protect\mathpalette{\protect\independenT}{\perp}}
\def\independenT#1#2{\mathrel{\rlap{$#1#2$}\mkern2mu{#1#2}}}

\usepackage{hyperref}
\newcommand{\sag}[2]{\tikz{\node[shape=circle,draw,inner sep=1pt,minimum width = 0.6cm, fill=#1]{$X_{#2}$};}} 
\newcommand{\blind}{1}

\begin{document}

\def\spacingset#1{\renewcommand{\baselinestretch}%
{#1}\small\normalsize} \spacingset{1}

\if1\blind
{
 \title{\bf Learning and interpreting asymmetry-labeled DAGs: a case study on COVID-19 fear}
  \author{ 
Manuele Leonelli \\
School of Science and Technology, IE University, Madrid, Spain\\
and\\
Gherardo Varando\\
Image Processing Laboratory, Universitat de València,  València, Spain}
  \maketitle
} \fi

\if0\blind
{
  \bigskip
  \bigskip
  \bigskip
  \begin{center}
    {\LARGE\bf Title}
\end{center}
  \medskip
} \fi

\bigskip
\begin{abstract}
Bayesian networks are widely used to learn and reason about the dependence structure of discrete variables. However, they are only capable of formally encoding symmetric conditional independence, which in practice is often too strict to hold. Asymmetry-labeled DAGs have been recently proposed to both extend the class of Bayesian networks by relaxing the symmetric assumption of independence and denote the type of dependence existing between the variables of interest. Here, we introduce novel structural learning algorithms for this class of models which, whilst being efficient, allow for a straightforward interpretation of the underlying dependence structure. A comprehensive computational study highlights the efficiency of the algorithms. A real-world data application using data from the Fear of COVID-19 Scale collected in Italy showcases their use in practice.
\end{abstract}

\noindent%

\spacingset{1.45} % DON'T change the spacing!

\section{Introduction}

Bayesian networks (BNs) are probabilistic graphical models which concisely represent the dependence structure between discrete variables through a directed acyclic graph (DAG) \cite{Pearl,Sucar}. Any conditional independence between variables embedded in the model can be directly read from the underlying DAG through the so-called D-separation criterion \cite{Pearl2009}. 

However, in practical applications, it has been found that often the symmetric assumption of conditional independence is too restrictive and models graphically depicting asymmetric independence are needed. Various notions of asymmetric conditional independence have been since defined, including context-specific \cite{Boutilier1996}, partial \cite{Pensar2016} and local \cite{Chickering1997}, and formal studies of their properties appeared \cite{Corander2019,Duarte2021,Shen2020,Tikka2019}. Although extensions of BNs embedding and representing asymmetric conditional independence have been defined \cite{Friedman1996,Jaeger2006,Pensar2015,Poole2003,Talvitie2019}, they often lose the intuitiveness associated to DAGs and no software is available for their use in practice.

Asymmetry-labeled DAGs (ALDAGs) have been recently introduced as an extension of DAGs where edges are labeled according to the type of dependence existing between any pairs of random variables \cite{Varando2022aldag}. They are constructed starting from a tree-based probabilistic graphical model called \emph{staged tree} \cite{Collazo2018,Smith2008}, to which a conversion algorithm is applied to construct the associated ALDAG. Standard tools for DAGs, e.g. the already-mentioned D-separation criterion, also hold for ALDAGs. However, they also carry additional information in the form of both the associated edge labeling and the underlying staged tree used to construct them. 

In \cite{leonelli2022highly} a structural learning algorithm for sparse ALDAGs, meaning with a small number of edges, is proposed. This consists of three steps: (i) a DAG with an upper-bound on the number of parents is learned from data; (ii) the learned DAG is refined into a staged tree embedding asymmetric conditional independences; (iii) the staged tree is converted into its associated ALDAG representation. However, this routine suffers from various drawbacks:
\begin{itemize}
    \item the ordering of the variables is chosen from one arbitrary topological order of the DAG in step (i), which may not be optimal for the ALDAG since it only considers symmetric forms of independence;
    \item the parents of a variable are also chosen from the DAG in step (i), and again this choice may be highly affected by overlooking asymmetric dependences.
\end{itemize}

Here we propose novel structural learning algorithms for ALDAGs which overcome these difficulties by replacing step (i) of learning a DAG from data with a procedure that both chooses the ordering of the variables and the parents of each variable in the final ALDAG. To this end, we take advantage of the notion of conditional mutual information already utilized in \cite{carli2022}, to quantify the strength of dependence between variables.

The algorithms that we introduce here learn sparse ALDAGs with a small number of connections between variables. Sparsity has two major advantages. First, the learning algorithms are more efficient since they search for an optimal model over a restricted number of available ALDAGs, speeding up computations. Second, they allow for straightforward and interpretable visualization of asymmetric dependence, which would not be feasible for generic ALDAGs. This is in line with the most recent trends in artificial intelligence and machine learning in focusing on being able to explain the outputs of a model \cite{XAI2,XAI1,XAI3}.

The paper is structured as follows. Section \ref{sec:aldag} reviews DAGs, ALDAGs, and various notions of both symmetric and asymmetric conditional independence. Section \ref{sec:learning} first reviews staged trees, which are needed for constructing ALDAGs, and then introduces novel structural learning routines for ALDAGs that are both efficient and explainable. Section \ref{sec:simulation} presents a simulation study to investigate the quality of our algorithms. Section \ref{sec:data} presents a comprehensive analysis of the use of ALDAGs to study the relationship between factors influencing the fear of the COVID-19 virus. The paper is concluded with a discussion.

\section{Asymmetric dependence and ALDAGs}
\label{sec:aldag}
\subsection{DAGs and conditional independence}

 Let $G=([p],F)$ be a directed acyclic graph (DAG) with vertex set $[p]=\{1,\dots,p\}$ and edge set $F$. Let $\bm{X}=(X_i)_{i\in[p]}$ be categorical random variables with joint mass function $P$ and sample space $\mathbb{X}=\times_{i\in[p]}\mathbb{X}_i$. For $A\subset [p]$, we let $\bm{X}_A=(X_i)_{i\in A}$ and $\bm{x}_A=(x_i)_{i\in A}$ where $\bm{x}_A\in\mathbb{X}_A=\times_{i\in A}\mathbb{X}_i$. We say that $P$ is Markov to $G$ if, for $\bm{x}\in\mathbb{X}$, 
\[
P(\bm{x})=\prod_{k\in[p]}P(x_k | \bm{x}_{\Pi_k}),
\]
where $\Pi_k$ is the parent set of $k$ in $G$ and $P(x_k | \bm{x}_{\Pi_k})$ is a shorthand for $P(X_k=x_k |\bm{X}_{\Pi_k} = \bm{x}_{\Pi_k})$. The ordered Markov condition implies conditional independences of the form
\begin{equation}
\label{ci}
X_i \independent \bm{X}_{[i-1]}\,|\, \bm{X}_{\Pi_i},
\end{equation}
which are equivalent to 
\begin{equation}
\label{ci2}
P(x_i|\bm{x}_{[i-1]\setminus \Pi_i}, \bm{x}_{\Pi_i})= P(x_i|\bm{x}_{\Pi_i}), \hspace{0.5cm} \mbox{for all }  \bm{x}\in\mathbb{X}.
\end{equation}

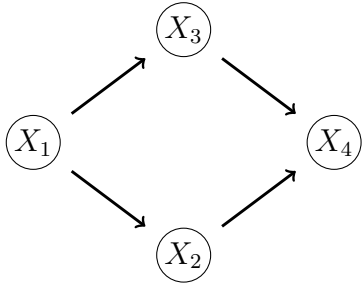
\begin{figure}
\centering
\begin{tikzpicture}
\renewcommand{\xx}{2}
\renewcommand{\yy}{1.5}
\node (1) at (0*\xx,1*\yy){\sag{white}{1}};
\node (2) at (1*\xx,0*\yy){\sag{white}{2}};
\node (3) at (1*\xx,2*\yy){\sag{white}{3}};
\node (4) at (2*\xx,1*\yy){\sag{white}{4}};
\draw[->, line width = 1.1pt] (1) -- (2);
\draw[->, line width = 1.1pt] (1) -- (3);
\draw[->, line width = 1.1pt] (3) -- (4);
\draw[->, line width = 1.1pt] (2) -- (4);
\end{tikzpicture}
\caption{An example of a DAG over four random variables. \label{fig:bn}}
\end{figure}

Figure \ref{fig:bn} shows a DAG over four random variables $X_1,\dots,X_4$, implying the symmetric conditional independences $X_3\independent X_2|X_1$ and $X_4\independent X_1| X_2,X_3$. The associated Markov factorization is $P(x_4|x_3,x_2)P(x_3|x_2)P(x_2|x_1)P(x_1)$.

\subsection{Asymmetric conditional independence}
BNs have the capability of expressing only symmetric conditional independence of the form in (\ref{ci}) and (\ref{ci2}). The most common asymmetric extension of conditional independence is the so-called \emph{context-specific} independence which is often represented by associating a tree to each vertex of a BN \cite{Boutilier1996}.  Let $A$, $B$ and $C$ be three disjoint subsets of $[p]$. We say that $\bm{X}_A$ is context-specific independent of $\bm{X}_B$ given context $\bm{x}_C\in\mathbb{X}_C$ if 
\begin{equation}
\label{eq:csi}
P(\bm{x}_A|\bm{x}_B,\bm{x}_C)=P(\bm{x}_A|\bm{x}_C)
\end{equation}
holds for all $(\bm{x}_A,\bm{x}_B)\in\mathbb{X}_{A\cup B}$ and write $\bm{X}_A\independent \bm{X}_B|\bm{x}_C$. The condition in (\ref{eq:csi}) reduces to standard conditional independence in (\ref{ci}) if it holds for all $\bm{x}_C\in\mathbb{X}_C$.

 A more general definition of non-symmetric conditional independence called \emph{partial conditional independence} was introduced in \cite{Pensar2016}. We say that $\bm{X}_A$ is partially conditionally independent of $\bm{X}_B$ in the domain $\mathcal{D}_B\subseteq \mathbb{X}_B$ given context $\bm{X}_C=\bm{x}_C$ if
\begin{equation}
\label{eq:pci}
P(\bm{x}_A|\bm{x}_B,\bm{x}_C)=P(\bm{x}_A|\tilde{\bm{x}}_B,\bm{x}_C)
\end{equation}
holds for all $(\bm{x}_A,\bm{x}_B),(\bm{x}_A,\tilde{\bm{x}}_B)\in\mathbb{X}_A\times \mathcal{D}_B$ and write $\bm{X}_A\independent \bm{X}_B|\mathcal{D}_B,\bm{x}_C$. Clearly, (\ref{eq:csi}) and $(\ref{eq:pci})$ coincide if $\mathcal{D}_B=\mathbb{X}_B$. Furthermore, the sample space $\mathbb{X}_B$ must contain more than two elements for a non-trivial partial conditional independence to hold.

A final condition is the so-called \emph{local conditional independence} as first discussed in \cite{Chickering1997}. For $i\in[p]$ and an $A\subset[p]$ such that $A\cap \{i\}=\emptyset$, local conditional independence expresses equalities of probabilities of the form
\begin{equation}
\label{eq:lci}
P(x_i|\bm{x}_A)= P(x_i|\tilde{\bm{x}}_A)
\end{equation}
for all $x_i\in\mathbb{X}_i$ and two $\bm{x}_A,\tilde{\bm{x}}_A\in\mathbb{X}_A$. Notice that in terms of generality, \ref{eq:csi} $\preceq$ \ref{eq:pci} $\preceq$ \ref{eq:lci}. Condition \ref{eq:lci} simply states that some conditional probability distributions are identical, where no discernable patterns as in Equations (\ref{eq:csi}) and (\ref{eq:pci}) can be detected.

\subsection{Classes of dependence}

Suppose that a DAG model is available, but that we believe additional structure which is not necessarily symmetric exists between the variables. Such a structure might consist of equalities as those in Equations \ref{eq:csi}-\ref{eq:lci} associated with asymmetric independences. The following definition formalizes the types of dependence that might exist between two random variables that are joined by an edge in a DAG $G$.

\begin{definition}
\label{def:class}
Let  $P$ be  the joint mass function of $\mathbf{X}$ and  $P$ be Markov for a DAG $G= ([p], F)$. For each $(j,i) \in F$ we say that the dependence of $X_i$ from $X_j$ is of class
\begin{itemize}
\item \emph{context}, if $X_i$ and $X_j$ are context-specific independent given 
	some context $\mathbf{x}_{C}$ with $C =  \Pi_i \setminus \{j\}$. 
\item \emph{partial}, if $X_i$ is partially conditionally independent of $X_j$ in a
	domain $\mathcal{D}_j \subset \mathbb{X}_j$ given 
	a context $\mathbf{x}_{C}$ with $C =  \Pi_i \setminus \{j\}$; and
		$X_i$ and $X_j$ are not context-specific independent given the same context 
		$\mathbf{x}_C$. 
	\item \emph{local}, if none of the above hold and a local independence of the 
		form $P(x_i| \mathbf{x}_{\Pi_i}) = P(x_i| \tilde{\mathbf{x}}_{\Pi_i})$ is valid 
		where $x_j \neq \tilde{x}_j$.
\item \emph{total}, if none of the above hold. 
\end{itemize}
\end{definition}

In standard DAG models, all edges have the total label, but additional labels might be considered to denote more flexible types of dependence. Notice that if the class of dependence between $X_i$ and $X_j$ is context or
partial then there may also be  local independence statements as in Equation
\ref{eq:lci} involving these two variables. 
Similarly, the dependence between $X_i$ and $X_j$ can be both context and
partial concerning two different contexts. 
On the other hand, if their class
of dependence is local then, by definition,
there are no context-specific or partial equalities. The lack of an edge is associated with conditional independence as formalized in Equation \ref{ci} and as in standard DAGs.

\subsection{ALDAGs}
Given the classes of dependence introduced in Definition \ref{def:class}, we can now embellish a DAG with additional information about asymmetric dependence structure. The resulting labeled graph is called ALDAG in \cite{Varando2022aldag}, where
edges are colored depending on the label and therefore depending on the type of relationship between variables. 
 
Formally, let $G$ be a DAG, $F$ its edge set, and
\[\mathcal{L}^A=\{\textnormal{`context'}, \textnormal{`partial'}, \textnormal{`context/partial'}, \textnormal{`local'}, \textnormal{`total'} \}\] 
be the set of edge labels marking the type of dependence.

\begin{definition}
	An ALDAG is a pair $(G,\psi)$ where $G=([p], F)$ is a DAG and $\psi$ is a function from the edge set of $G$ to $\mathcal{L}^A$, i.e. $\psi: F\rightarrow \mathcal{L}^A$. We say that a joint mass function $P$ is compatible with an ALDAG 
	$(G, \psi)$ if $P$ is Markov to $G$ and additionally 
	$P$ respects all the edge labels given by $\psi$; that is,
	for each $(j,i) \in F$, $X_i$ is 
	$\psi(i,j)$ dependent from $X_j$.   
\end{definition}

Henceforth, we represent the labeling via a coloring of the edges of the ALDAG. Standard BNs have an ALDAG representation where all edges have the label `total'. Notice that standard features of BNs are also valid over ALDAGs: for instance, the already-mentioned d-separation criterion as well as fast probability propagation algorithms. Furthermore, they have been recently used for causal discovery by using the extra information given by the edge labels \cite{leonelli2022causal}.

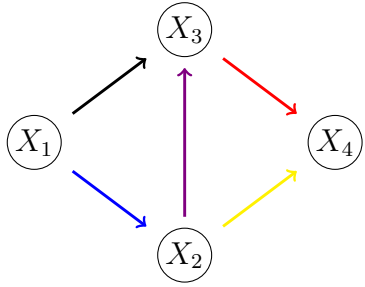
\begin{figure}
\centering
\begin{tikzpicture}
\renewcommand{\xx}{2}
\renewcommand{\yy}{1.5}
\node (1) at (0*\xx,1*\yy){\sag{white}{1}};
\node (2) at (1*\xx,0*\yy){\sag{white}{2}};
\node (3) at (1*\xx,2*\yy){\sag{white}{3}};
\node (4) at (2*\xx,1*\yy){\sag{white}{4}};
\draw[->,  blue, line width = 1.1pt] (1) -- (2);
\draw[->, black, line width = 1.1pt] (1) -- (3);
\draw[->, red, line width = 1.1pt] (3) -- (4);
\draw[->, yellow, line width = 1.1pt] (2) -- (4);
\draw[->, violet, line width = 1.1pt] (2) -- (3);

\end{tikzpicture}
\caption{An example of an ALDAG over four random variables. The edge coloring is: red - context; blue - partial; violet - context/partial; green - local; black - total.\label{fig:aldag}}
\end{figure}

Figure \ref{fig:aldag} gives an example of an ALDAG over four random variables. The lack of the edge from $X_1$ to $X_4$ is associated with the symmetric conditional independence $X_4\independent X_1|X_2,X_3$. The black edge with label total represents the fact that the relationship between $X_1$ and $X_3$ is symmetric. All other pairs of variables connected by an edge, which do not have a label total, highlight the fact that there are some equalities in the conditional probabilities of the model which are associated with asymmetric independences. Furthermore, in the case of the edge between $X_2$ and $X_4$ which has a label local, this asymmetric dependence is very unstructured and cannot be formalized either as context-specific or as partial.  

ALDAGs share features with labeled DAGs of \cite{Pensar2015} but they differ in two critical aspects: first, labeled DAGs can only embed context-specific independence whilst ALDAGs represent any type of asymmetric independence; second, labeled DAGs specifically report the contexts over which independences hold, whilst ALDAGs do not. There are two reasons behind this: on one hand, the specific independences in ALDAGs can be read from the associated staged tree (see below); on the other, for applications with a larger number of variables the required contexts are often too complex to be reported within the DAG.

\section{Learning ALDAGs from data}
\label{sec:learning}

Given observations from a vector of categorical variables, we want to use structural learning algorithms to learn the ALDAG that better describes the data dependence structure. Such algorithms will use the class of staged tree graphical models \cite{Collazo2018,Smith2008}, which are henceforth reviewed next.

\subsection{Staged trees}

Different from BNs, whose graphical representation is a DAG, staged trees visualize conditional independence using a colored tree. Let $(V,E)$ be a directed, finite, rooted tree with vertex set $V$, root node $v_0$, and edge set $E$. 
For each $v\in V$, 
let $E(v)=\{(v,w)\in E\}$ be the set of edges emanating
from $v$ and $\mathcal{C}$ be a set of labels. 
%Given a vector $\bm{x}$, we denote with $\bm{x}_{-i}$ the vectorwith\bm{x}$ without its $i$-th entry.

An $\bf X$-compatible staged tree %to a categorical random vector $\bm{X}$
is a triple $T = (V,E,\theta)$, where $(V,E)$ is a rooted directed tree and:
\begin{enumerate}
    \item $V = {v_0} \cup \bigcup_{i \in [p]} \mathbb{X}_{[i]}$;
		\item For all $v,w\in V$,
$(v,w)\in E$ if and only if $w=\bm{x}_{[i]}\in\mathbb{X}_{[i]}$ and 
			$v = \bm{x}_{[i-1]}$, or $v=v_0$ and $w=x_1$ for some
$x_1\in\mathbb{X}_1$;
\item $\theta:E\rightarrow \mathcal{L}=\mathcal{C}\times \cup_{i\in[p]}\mathbb{X}_i$ is a labelling of the edges such that $\theta(v,\bm{x}_{[i]}) = (\kappa(v), x_i)$ for some 
			function $\kappa: V \to \mathcal{C}$. The function 
			$k$ is called the \textit{colouring} of the staged tree $T$.
%\item $\kappa: V \to \mathcal{C}$ is a colouring of the vertices.
\end{enumerate}
	If $\theta(E(v)) = \theta(E(w))$ then $v$ and $w$ are said to be in the same 	\emph{stage}. The equivalence classes induced by  $\theta(E(v))$
form a partition of the internal vertices of the tree in \emph{stages}.

Points 1 and 2 above construct a rooted tree where each root-to-leaf path, or equivalently each leaf, is associated with an element of the sample space $\mathbb{X}$.  Then a labeling of the edges of such a tree is defined where labels are pairs with one element from a set $\mathcal{C}$ and the other from the sample space $\mathbb{X}_i$ of the corresponding variable $X_i$ in the tree. By construction, $\bf X$-compatible staged trees are such that two vertices can be in the same stage if and only if they correspond to the same sample space.

\begin{figure}
\centering
\scalebox{0.7}{
\begin{tikzpicture}
\renewcommand{\xx}{2}
\renewcommand{\yy}{1.05}
\node (v0) at (0*\xx,0*\yy) {\sage{white}{0}};
\node (v1) at (1*\xx,-6*\yy) {\sage{cyan}{1}};
\node (v2) at (1*\xx,-0*\yy) {\sage{cyan}{2}};
\node (v3) at (1*\xx,6*\yy) {\sage{yellow}{3}};
\node (v4) at (2*\xx,-8*\yy) {\sage{red}{4}};
\node (v5) at (2*\xx,-6*\yy) {\sage{red}{5}};
\node (v6) at (2*\xx,-4*\yy) {\sage{red}{6}};
\node (v7) at (2*\xx,-2*\yy) {\sage{green}{7}};
\node (v8) at (2*\xx,0*\yy) {\sage{green}{8}};
\node (v9) at (2*\xx,2*\yy) {\sage{purple}{9}};
\node (v10) at (2*\xx,4*\yy) {\sage{teal}{10}};
\node (v11) at (2*\xx,6*\yy) {\sage{pink}{11}};
\node (v12) at (2*\xx,8*\yy) {\sage{lightgray}{12}};
\node (v13) at (3*\xx,-8.5*\yy) {\sage{blue}{13}};
\node (v14) at (3*\xx,-7.5*\yy) {\sage{blue}{14}};
\node (v15) at (3*\xx,-6.5*\yy) {\sage{orange}{15}};
\node (v16) at (3*\xx,-5.5*\yy) {\sage{magenta}{16}};
\node (v17) at (3*\xx,-4.5*\yy) {\sage{brown}{17}};
\node (v18) at (3*\xx,-3.5*\yy) {\sage{orange}{18}};
\node (v19) at (3*\xx,-2.5*\yy) {\sage{blue}{19}};
\node (v20) at (3*\xx,-1.5*\yy) {\sage{blue}{20}};
\node (v21) at (3*\xx,-0.5*\yy) {\sage{orange}{21}};
\node (v22) at (3*\xx,0.5*\yy) {\sage{magenta}{22}};
\node (v23) at (3*\xx,1.5*\yy) {\sage{brown}{23}};
\node (v24) at (3*\xx,2.5*\yy) {\sage{orange}{24}};
\node (v25) at (3*\xx,3.5*\yy) {\sage{blue}{25}};
\node (v26) at (3*\xx,4.5*\yy) {\sage{blue}{26}};
\node (v27) at (3*\xx,5.5*\yy) {\sage{orange}{27}};
\node (v28) at (3*\xx,6.5*\yy) {\sage{magenta}{28}};
\node (v29) at (3*\xx,7.5*\yy) {\sage{brown}{29}};
\node (v30) at (3*\xx,8.5*\yy) {\sage{orange}{30}};
\node (v31) at (4*\xx,-8.75*\yy) {\leaf};
\node (v32) at (4*\xx,-8.25*\yy) {\leaf};
\node (v33) at (4*\xx,-7.75*\yy) {\leaf};
\node (v34) at (4*\xx,-7.25*\yy) {\leaf};
\node (v35) at (4*\xx,-6.75*\yy) {\leaf};
\node (v36) at (4*\xx,-6.25*\yy) {\leaf};
\node (v37) at (4*\xx,-5.75*\yy) {\leaf};
\node (v38) at (4*\xx,-5.25*\yy) {\leaf};
\node (v39) at (4*\xx,-4.75*\yy) {\leaf};
\node (v40) at (4*\xx,-4.25*\yy) {\leaf};
\node (v41) at (4*\xx,-3.75*\yy) {\leaf};
\node (v42) at (4*\xx,-3.25*\yy) {\leaf};
\node (v43) at (4*\xx,-2.75*\yy) {\leaf};
\node (v44) at (4*\xx,-2.25*\yy) {\leaf};
\node (v45) at (4*\xx,-1.75*\yy) {\leaf};
\node (v46) at (4*\xx,-1.25*\yy) {\leaf};
\node (v47) at (4*\xx,-0.75*\yy) {\leaf};
\node (v48) at (4*\xx,-0.25*\yy) {\leaf};
\node (v49) at (4*\xx,0.25*\yy) {\leaf};
\node (v50) at (4*\xx,0.75*\yy) {\leaf};
\node (v51) at (4*\xx,1.25*\yy) {\leaf};
\node (v52) at (4*\xx,1.75*\yy) {\leaf};
\node (v53) at (4*\xx,2.25*\yy) {\leaf};
\node (v54) at (4*\xx,2.75*\yy) {\leaf};
\node (v55) at (4*\xx,3.25*\yy) {\leaf};
\node (v56) at (4*\xx,3.75*\yy) {\leaf};
\node (v57) at (4*\xx,4.25*\yy) {\leaf};
\node (v58) at (4*\xx,4.75*\yy) {\leaf};
\node (v59) at (4*\xx,5.25*\yy) {\leaf};
\node (v60) at (4*\xx,5.75*\yy) {\leaf};
\node (v61) at (4*\xx,6.25*\yy) {\leaf};
\node (v62) at (4*\xx,6.75*\yy) {\leaf};
\node (v63) at (4*\xx,7.25*\yy) {\leaf};
\node (v64) at (4*\xx,7.75*\yy) {\leaf};
\node (v65) at (4*\xx,8.25*\yy) {\leaf};
\node (v66) at (4*\xx,8.75*\yy) {\leaf};
\draw[->] (v0) -- node [below, sloped] {\tiny{low}} (v1);
\draw[->] (v0) -- node [below, sloped] {\tiny{medium}} (v2);
\draw[->] (v0) --  node [above, sloped] {\tiny{high}} (v3);
\draw[->] (v1) --  node [below, sloped] {\tiny{low}} (v4);
\draw[->] (v1) --  node [above, sloped] {\tiny{medium}} (v5);
\draw[->] (v1) --  node [above, sloped] {\tiny{high}} (v6);
\draw[->] (v2) --  node [below, sloped] {\tiny{low}} (v7);
\draw[->] (v2) --  node [above, sloped] {\tiny{medium}} (v8);
\draw[->] (v2) --  node [above, sloped] {\tiny{high}} (v9);
\draw[->] (v3) --  node [below, sloped] {\tiny{low}} (v10);
\draw[->] (v3) --  node [above, sloped] {\tiny{medium}} (v11);
\draw[->] (v3) --  node [above, sloped] {\tiny{high}} (v12);
\draw[->] (v4) --  node [below, sloped] {\tiny{low}} (v13);
\draw[->] (v4) --  node [above, sloped] {\tiny{high}} (v14);
\draw[->] (v5) --  node [below, sloped] {\tiny{low}} (v15);
\draw[->] (v5) --  node [above, sloped] {\tiny{high}} (v16);
\draw[->] (v6) --  node [below, sloped] {\tiny{low}} (v17);
\draw[->] (v6) --  node [above, sloped] {\tiny{high}} (v18);
\draw[->] (v7) --  node [below, sloped] {\tiny{low}} (v19);
\draw[->] (v7) --  node [above, sloped] {\tiny{high}} (v20);
\draw[->] (v8) --  node [below, sloped] {\tiny{low}} (v21);
\draw[->] (v8) --  node [above, sloped] {\tiny{high}} (v22);
\draw[->] (v9) --  node [below, sloped] {\tiny{low}} (v23);
\draw[->] (v9) --  node [above, sloped] {\tiny{high}} (v24);
\draw[->] (v10) --  node [below, sloped] {\tiny{low}} (v25);
\draw[->] (v10) --  node [above, sloped] {\tiny{high}} (v26);
\draw[->] (v11) --  node [below, sloped] {\tiny{low}} (v27);
\draw[->] (v11) --  node [above, sloped] {\tiny{high}} (v28);
\draw[->] (v12) --  node [below, sloped] {\tiny{low}} (v29);
\draw[->] (v12) --  node [above, sloped] {\tiny{high}} (v30);
\draw[->] (v13) --  node [below, sloped] {\tiny{low}} (v31);
\draw[->] (v13) --  node [above, sloped] {\tiny{high}} (v32);
\draw[->] (v14) --  node [below, sloped] {\tiny{low}} (v33);
\draw[->] (v14) --  node [above, sloped] {\tiny{high}} (v34);
\draw[->] (v15) --  node [below, sloped] {\tiny{low}} (v35);
\draw[->] (v15) --  node [above, sloped] {\tiny{high}} (v36);
\draw[->] (v16) --  node [below, sloped] {\tiny{low}} (v37);
\draw[->] (v16) --  node [above, sloped] {\tiny{high}} (v38);
\draw[->] (v17) --  node [below, sloped] {\tiny{low}} (v39);
\draw[->] (v17) --  node [above, sloped] {\tiny{high}} (v40);
\draw[->] (v18) --  node [below, sloped] {\tiny{low}} (v41);
\draw[->] (v18) --  node [above, sloped] {\tiny{high}} (v42);
\draw[->] (v19) --  node [below, sloped] {\tiny{low}} (v43);
\draw[->] (v19) --  node [above, sloped] {\tiny{high}} (v44);
\draw[->] (v20) --  node [below, sloped] {\tiny{low}} (v45);
\draw[->] (v20) --  node [above, sloped] {\tiny{high}} (v46);
\draw[->] (v21) --  node [below, sloped] {\tiny{low}} (v47);
\draw[->] (v21) --  node [above, sloped] {\tiny{high}} (v48);
\draw[->] (v22) --  node [below, sloped] {\tiny{low}} (v49);
\draw[->] (v22) --  node [above, sloped] {\tiny{high}} (v50);
\draw[->] (v23) --  node [below, sloped] {\tiny{low}} (v51);
\draw[->] (v23) --  node [above, sloped] {\tiny{high}} (v52);
\draw[->] (v24) --  node [below, sloped] {\tiny{low}} (v53);
\draw[->] (v24) --  node [above, sloped] {\tiny{high}} (v54);
\draw[->] (v25) --  node [below, sloped] {\tiny{low}} (v55);
\draw[->] (v25) --  node [above, sloped] {\tiny{high}} (v56);
\draw[->] (v26) --  node [below, sloped] {\tiny{low}} (v57);
\draw[->] (v26) --  node [above, sloped] {\tiny{high}} (v58);
\draw[->] (v27) --  node [below, sloped] {\tiny{low}} (v59);
\draw[->] (v27) --  node [above, sloped] {\tiny{high}} (v60);
\draw[->] (v28) --  node [below, sloped] {\tiny{low}} (v61);
\draw[->] (v28) --  node [above, sloped] {\tiny{high}} (v62);
\draw[->] (v29) --  node [below, sloped] {\tiny{low}} (v63);
\draw[->] (v29) --  node [above, sloped] {\tiny{high}} (v64);
\draw[->] (v30) --  node [below, sloped] {\tiny{low}} (v65);
\draw[->] (v30) --  node [above, sloped] {\tiny{high}} (v66);
\end{tikzpicture}
}

\caption{A staged tree over two ternary variables and two binary variables, whose ALDAG is in Figure \ref{fig:aldag}. \label{fig:staged}}
\end{figure}
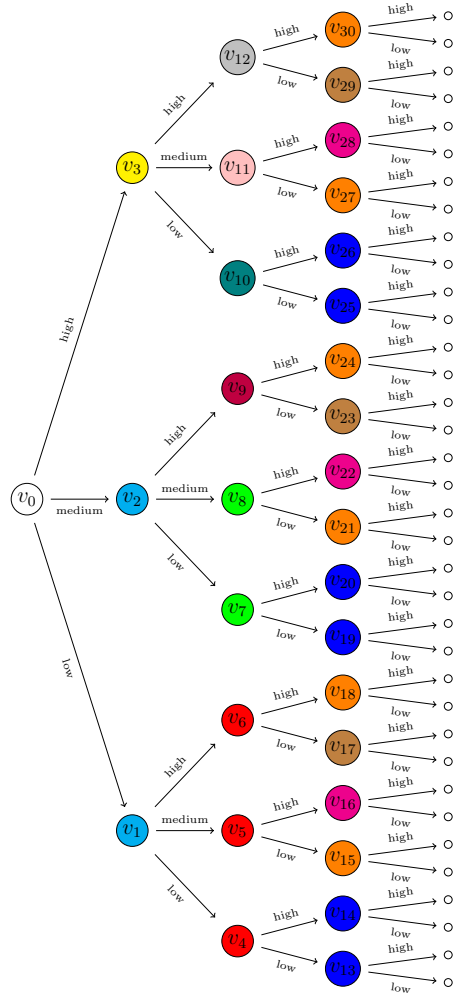

Figure \ref{fig:staged} reports an example of an $(X_1,X_2,X_3,X_4)$-compatible staged tree model over two ternary variables ($X_1$ and $X_2$ with levels low/medium/high) and two binary ones ($X_3$ and $X_4$ with levels low/high). The coloring given by the function $\kappa$ is shown in the vertices and each edge $(\cdot, (x_1,\dots, x_i))$ is labeled with $x_i$. The edge labeling $\theta$ can be read from the graph by combining the text label and the color of the emanating vertex.  For example, $ \theta(v_1, v_6) \neq \theta(v_2, v_7) 
= \theta(v_2,v_{8}) \neq \theta(v_2, v_{9})$.  This representation of the labeling $\theta$  over vertices is equivalent to that over edges, whilst being more interpretable, and is henceforth used. There are 31 internal vertices and the staging is $\{v_0\}$, $\{v_1,v_2\}$, $\{v_3\}$,
$\{v_4,v_5,v_6\}$, $\{v_7,v_{8}\}$,  $\{v_9\}$, $\{v_{10}\}$, $\{v_{11}\}$,
$\{v_{12}\}$, $\{v_{13},v_{14},v_{19},v_{20},v_{25},v_{26}\}$, $\{v_{15},v_{18},v_{21},v_{24},v_{27},v_{30}\}$, $\{v_{16},v_{22},v_{28}\}$ and $\{v_{17},v_{23},v_{29}\}$.

Conditional independence is formally modeled and represented in staged trees via the labeling $\theta$. As an illustration, consider the staged tree in Figure \ref{fig:staged}. The fact that $v_4, v_5$ and $v_6$ are in the same stage is associated with the context-specific independence $X_3\independent X_2| X_1 = \textnormal{low}$. The green stage $\{v_7,v_8\}$ represents the partial independence between $X_2$ and $X_3$ in the domain $X_2\in\{\textnormal{low,medium}\}$ and context $X_1=\textnormal{medium}$. The fact that, for instance, $v_{27}$ and $v_{30}$ are in the same stage is a generic local independence stating that the conditional distribution of $X_4$ given $X_1=\textnormal{high},X_2=\textnormal{high},X_3=\textnormal{high}$ is equal to that of $X_4$ given $X_1=\textnormal{high},X_2=\textnormal{medium},X_3=\textnormal{low}$. Lastly, the repeating staging pattern in $v_{13}-v_{18}$, $v_{19}-v_{24}$ and $v_{25}-v_{30}$ represents the symmetric conditional independence $X_4\independent X_1| X_2,X_3$. 

Staged trees represent a wide array of asymmetric independences via the staging of the internal vertices. However, as the number of possible atomic events increases, it becomes more challenging to assess the underlying independences by visual inspection. For this reason, \cite{Varando2022aldag} introduced ALDAGs as a compression of the information stored in a staged tree which can be more intuitively read, as well as an algorithm to convert a staged tree into its ALDAG representation. Such a conversion algorithm requires two steps: (i) constructing a DAG $G_T$ from the staged tree $T$ so that if $\bm{X}_A\independent \bm{X}_B |\bm{X}_C$ in $T$ then $\bm{X}_A$ and $\bm{X}_B$ are d-separated by $\bm{X}_C$ in $G_T$; (ii) labeling each edge in $G_T$ according to the type of dependence existing between the associated variables. As an illustration, the ALDAG associated with the staged tree in Figure \ref{fig:staged} is the one in Figure \ref{fig:aldag}. The lack of the edge ($X_1$, $X_4$) follows from the repeating staging pattern in  $v_{13}-v_{18}$, $v_{19}-v_{24}$ and $v_{25}-v_{30}$. The labeled edges give a summary of the very complex staging of the staged tree, which in general cannot be simply represented by a standard BN.

\subsection{Learning ALDAGs with a fixed order}

Given the conversion algorithm from staged tree to ALDAG introduced in \cite{Varando2022aldag}, learning an ALDAG from data comes down to learning the associated staged tree. There is now a wide array of algorithms to learn different types of staged trees \cite{barclay,collazo,freeman,leonelli2022structural,leonelli2022highly,silander}, which are also implemented in the freely-available \texttt{stagedtrees} R package \cite{carliR}.  Because the model search space of staged trees is huge and much larger than that of DAGs, all these algorithms are based on greedy heuristic searches that at each iteration optimize a model score (most often the BIC as discussed in \cite{gorgen2022}). 

Research has been pursued in simplifying the task of learning a staged tree by considering only specific sub-classes.  In \cite{carli2022}  naive staged trees are defined which have the same number of parameters of a naive BN over
the same variables. In \cite{leonelli2022structural} simple staged trees are considered which have
a constrained type of partitioning of the vertices. In \cite{Duarte2021} CStrees are defined
which only embed symmetric and context-specific types of independence. In \cite{Varando2022aldag} algorithms that learn trees whose ALDAG does not have local edges are studied. Lastly, in \cite{leonelli2022highly} k-parents staged trees are defined which have ALDAGs with a limited number of parents. Since these are used in our novel algorithms for sparse ALDAGs, we recall here their formal definition.

\begin{definition}
A staged tree $T$ is in the class of $k$-parents staged trees if the maximum in-degree in $G_T$ is less or equal to $k$.
\end{definition}

One of the first solutions to make structural learning of BNs scalable was to limit the number of parents each variable can have \cite{friedman,tsar}. This was imposed not only to restrict the model space of possible DAGs but  also made sense from an applied point of view since most often only a limited number of variables can be expected to directly influence another. The option of setting a maximum number of parents is also available in the standard \texttt{bnlearn} software \cite{scutari}.

\subsection{Learning sparse ALDAGs with a fixed order}
\label{sec:fixed}

The ALDAG associated with a $k$-parents staged tree is consequently sparse whenever $k$ is set to a small integer. Notice that in \cite{collazo} staged trees whose ALDAGs would be sparse are learned using non-local priors, which have been also effective in learning sparse DAGs \cite{altomare}. 

Here we consider more flexible alternatives to the algorithms discussed in \cite{leonelli2022highly}, which required utilizing a specific compatible order to a BN learned from data.  Let's also start by assuming that we select a priori a possible ordering of the variables of interest $X$ (henceforth we assume this is the ordering of the positive integers). To select which variables can be parents of a vertex in the learned ALDAG we use \textit{conditional mutual information}. Recall that the conditional mutual information between two variables $X_A$ and $X_B$ given (a vector) $\bm{X}_C$, $I(X_A;X_B|\bm{X}_C)$, is defined as 
\begin{equation}
\label{eq:cmi}
I(X_A;X_B|\bm{X}_C)=\sum_{\bm{x}_c\in\mathbb{X}_C}P(\bm{x}_C)\sum_{x_A\in\mathbb{X}_A}\sum_{x_B\in\mathbb{X}_C}P(x_A,x_B|\bm{x}_C)\ln\left(\frac{P(x_A,x_B|\bm{x}_C)}{P(x_A|\bm{x}_C)P(x_B|\bm{x}_C)}\right)
\end{equation}
In particular, the parents of a variable $X_i$ are selected iteratively by choosing the variable $X_j$ in $X_{[i-1]}$ maximizing the conditional mutual information with $X_i$ given the already selected parent variables. Of course, if $i\leq k+1$, then all preceding variables are parents of $X_i$. Notice that the probability distributions in Equation \ref{eq:cmi} need to be computed empirically from data: for this task, we use the \texttt{infotheo} R package \cite{infotheo}.

The previous routine constructs a DAG without estimating any of the associated probabilities. The structural learning algorithm for the ALDAG then takes this DAG as input and performs the following steps: (i) convert the DAG into an equivalent staged tree representation (embedding all the DAG conditional independences); (ii) run the backward-hill climbing algorithm of \cite{carliR}, which at each iteration can only join stages of the underlying staged tree; (iii) transform the resulting staged tree into $G_T$ and add the edge labels. By construction, the resulting ALDAG is such that any vertex has at most $k$ parents.

\subsection{Learning sparse ALDAGs without a fixed order}

The methods described in the previous section
output an $\bm{X}_{\pi}$-compatible staged tree for a possible ordering $\pi$ of the variables.  
For a small number of variables, it is possible to 
simply enumerate all  $p!$  possible orders, and select the best one(s) according to a chosen criterion  (e.g. BIC). We investigate below the feasibility of such an approach.

\subsection{Learning sparse ALDAGs with a partial order}
\label{sec:compatible}
An intermediate solution, to avoid the factorial increase in complexity of considering all possible orders, is to consider only those orders that are compatible with a learned BN from data. For each of the allowed orders, an ALDAG is learned from data using the algorithm of Section \ref{sec:fixed}, and the best-scoring one according to a chosen criterion (e.g. BIC) is selected.  There are three different possible ways to restrict the number of orders that we pursue here:

\begin{enumerate}
    \item Learn a BN from data using the tabu algorithm and select all compatible orders to the associated DAG;
    \item Learn a BN from data using the tabu algorithm and construct the mixed graph representing its equivalence class; we then select all compatible orders by considering the DAG consisting of only the directed edges of this mixed graph;
    \item Learn a mixed graph using the PC stable algorithm \cite{Colombo2014} which represents again an equivalence class;  we then select all compatible orders by considering the DAG consisting of only the directed edges of this mixed graph.
\end{enumerate}

The last two approaches consider possible orders where only causal relationships between variables \cite{Pearl2009} are fixed. By construction, the number of possible orders considered by option 2 cannot be smaller than option 1. 

Notice that the above algorithms require the learning of a BN, or a BN equivalence class, from data, but only to restrict the possible orders to be considered. The parents of the learned ALDAG do not depend on those of this initial BN and are chosen using conditional mutual information as in Section \ref{sec:fixed}.

\section{Experiments}
\label{sec:simulation}
\subsection{Computational study}

\begin{table}
\centering
\caption{BIC for models learned over nine datasets: 
                    Bayesian networks (DAG), 
                    ALDAGs with the approach of \cite{leonelli2022highly} (LV), 
                    ALDAGs with a fixed order (CMI), 
                    ALDAGs using compatible orders (ORD1, ORD2 and ORD3), 
                    ALDAGs considering all orders (ALL).} 
\scalebox{0.6}{
\begin{tabular}{llccccccc}
  \toprule
 data & k & DAG & LV & CMI & ORD1 & ORD2 & ORD3 & ALL \\ 
  \midrule
 \multirow{3}{4em}{asia} &   1 & \textbf{22694.01} & \textbf{22694.01} & 22919.69 & 22919.69 & 22919.69 & 22908.34 &  \\ 
&   2 & 22214.59 & 22190.48 & \textbf{22184.01} & \textbf{22184.01} & \textbf{22184.01} & \textbf{22184.01} &  \\ 
&   3 & 22214.59 & 22190.48 & 22181.24 & \textbf{22178.24} & \textbf{22178.24} & \textbf{22178.24} &  \\ 
 \midrule
 \multirow{3}{4em}{cachexia} &   1 & 963.98 & 958.85 & 973.35 & 958.85 & \textbf{957.14} & 973.12 &  \\ 
 &   2 & 963.98 & 958.85 & 925.16 & 920.22 & \textbf{904.47} & 914.28 &  \\ 
 &   3 & 963.98 & 958.85 & 877.76 & 871.25 & \textbf{851.55} & 873.27 &  \\ 
 \midrule
 \multirow{3}{4em}{chds} &   1 & 2831.56 & 2825.67 & 2825.67 & 2825.67 & \textbf{2825.61} & 2849.18 & \textbf{2825.61} \\ 
 &   2 & 2831.56 & 2825.67 & 2826.09 & 2825.67 & \textbf{2820.87} & \textbf{2820.87} & \textbf{2820.87} \\ 
&   3 & 2831.56 & 2825.67 & 2824.87 & 2823.81 & \textbf{2817.55} & \textbf{2817.55} & \textbf{2817.55} \\ 
 \midrule
 \multirow{3}{4em}{coronary} &   1 & \textbf{13507.86} & \textbf{13507.86} & 13508.53 & 13556.92 & 13555.14 & 13560.47 & \textbf{13507.86} \\ 
  &   2 & 13449.09 & 13425.35 & 13424.10 & 13490.45 & 13487.30 & 13500.24 & \textbf{13419.67} \\ 
  &   3 & 13442.02 & 13403.86 & 13385.54 & 13414.38 & 13411.23 & 13475.69 & \textbf{13370.25} \\ 
 \midrule
\multirow{3}{4em}{fall} &   1 & 138118.14 & \textbf{138060.31} & 142495.58 & \textbf{138060.31} & \textbf{138060.31} & \textbf{138060.31} & \textbf{138060.31} \\ 
  &   2 & 137637.86 & \textbf{137418.54} & 137561.62 & \textbf{137418.54} & \textbf{137418.54} & \textbf{137418.54} &\textbf{137418.54} \\ 
 &   3 & 137637.86 & \textbf{137418.54} & 137503.24 & \textbf{137418.54} & \textbf{137418.54} & \textbf{137418.54} & \textbf{137418.54} \\ 
 \midrule
 \multirow{3}{4em}{ksl} &   1 & \textbf{11927.73} & \textbf{11927.73} & 12197.26 & 12115.97 & 12115.97 & 12186.57 &  \\ 
  &   2 & 11801.14 & 11773.84 & 11960.43 & \textbf{11768.09} & \textbf{11768.09} & 11948.90 &  \\ 
 &   3 & 11801.14 & 11773.84 & 11940.19 & \textbf{11750.37} & \textbf{11750.37} & 11916.28 &  \\ 
 \midrule
 \multirow{3}{5em}{mathmarks} &   1 & 956.03 & 942.13 & 942.13 & 942.13 & \textbf{936.78} & 952.71 & \textbf{936.78} \\ 
 &   2 & 956.03 & 942.13 & 930.64 & 930.64 & \textbf{913.23} & 918.01 & \textbf{913.23} \\ 
 &   3 & 956.03 & 942.13 & 890.83 & 890.83 & \textbf{876.21} & 878.30 & \textbf{876.21} \\ 
\midrule
 \multirow{3}{4em}{phd} &   1 & 8372.67 & 8371.22 & \textbf{8369.02} & 8397.87 & 8397.87 & 8402.31 & \textbf{8369.02} \\ 
  &   2 & 8371.68 & 8356.90 & 8359.82 & \textbf{8345.74} & \textbf{8345.74} & 8348.98 & \textbf{8345.74} \\ 
 &   3 & 8371.68 & 8356.90 & 8339.95 & 8338.37 & 8338.37 & 8338.44 & \textbf{8330.42} \\ 
 \midrule
 \multirow{3}{4em}{titanic} &   1 & 10651.36 & \textbf{10641.16} & \textbf{10641.16} & \textbf{10641.16} & \textbf{10641.16} & \textbf{10641.16} & \textbf{10641.16} \\ 
  &   2 & 10502.28 & 10451.72 & 10499.21 & 10451.72 & \textbf{10450.72} & \textbf{10450.72} & \textbf{10450.72} \\ 
 &   3 & 10502.28 & 10451.72 & 10443.48 & 10432.84 & \textbf{10431.85} & \textbf{10431.85} & \textbf{10431.85} \\ 
   \bottomrule
\end{tabular}
}
\label{table:bic}
\end{table}

We start by considering nine datasets from the literature on probabilistic graphical models to assess the performance of sparse  ALDAGs and different estimation procedures. If required, variables are discretized using the equal-frequency method. For each dataset, six different
estimation procedures are considered and for each of these procedures, $k=1,2,3$ number of parents is used. First, a BN model is learned using the tabu algorithm
implemented in the bnlearn R package \cite{scutari} (labeled
DAG). A topological order of
the learned BN is then used to learn an ALDAG using the approach in \cite{leonelli2022highly} (labeled LV). The approaches proposed in this paper are then used: with a fixed order (labeled CMI) and using the three compatible order approaches of Section \ref{sec:compatible} (labeled ORD1, ORD2, and ORD3, respectively). Lastly, if computationally feasible, meaning when the number of variables is less than seven, we implemented an exhaustive model search for every possible variable ordering (labeled ALL).

The BIC of the learned models is reported in Table \ref{table:bic}. We can draw the following conclusions from the table:

\begin{itemize}
    \item For every dataset and every algorithm, the BIC decreases as the number of parents $k$ increases. 
    \item For the case $k=1$, the DAG approach outperforms some of the ALDAG algorithms, but for $k=2,3$ ALDAGs return lower BIC scores.
    \item In all cases, there is at least one algorithm that scores equal to or better than the standard DAG approach using tabu.
    \item Except for the coronary data and the phd data (with $k=3$), there is an ALDAG algorithm that scores equally well as the algorithm considering all possible orders.
    \item The learning algorithm starting from the equivalence class of the tabu DAG algorithm is overall the best-scoring one out of those that do not consider every possible order, and in particular, it outperforms the one starting from the output of the PC algorithm.
\end{itemize}

\begin{table}
\centering
\caption{Time (in seconds) to learn the models reported in Table \ref{table:bic}. Computations carried
out with an 11th Gen Intel(R) Core(TM) i7-1165G7 \@ 2.80GHz.
} 
\scalebox{0.6}{
\begin{tabular}{lcccccccc}
  \toprule
  data & k & DAG & LV& CMI & ORD1& ORD2 & ORD3 & ALL \\ 
  \midrule
 \multirow{3}{5em}{asia} &   1 & 0.02 & 0.07 & 0.16 & 34.72 & 169.97 & 1929.00 &  \\ 
 &   2 & 0.02 & 0.02 & 0.25 & 64.54 & 320.61 & 4530.72 &  \\ 
 &   3 & 0.02 & 0.03 & 0.36 & 102.28 & 526.64 & 7739.50 &  \\ 
 \midrule
 \multirow{3}{5em}{cachexia} &   1 & 0.00 & 0.01 & 0.03 & 0.44 & 151.39 & 19.50 &  \\ 
 &   2 & 0.00 & 0.03 & 0.08 & 0.95 & 390.86 & 50.76 &  \\ 
&   3 & 0.00 & 0.02 & 1.00 & 7.03 & 3237.02 & 376.25 &  \\ 
 \midrule
\multirow{3}{5em}{chds} &   1 & 0.01 & 0.00 & 0.00 & 0.02 & 0.14 & 0.06 & 0.12 \\ 
 &   2 & 0.00 & 0.00 & 0.02 & 0.03 & 0.24 & 0.07 & 0.22 \\ 
&   3 & 0.00 & 0.01 & 0.02 & 0.06 & 0.50 & 0.18 & 0.47 \\
 \midrule
 \multirow{3}{5em}{coronary} &   1 & 0.00 & 0.02 & 0.01 & 0.03 & 0.50 & 0.58 & 14.58 \\ 
 &   2 & 0.02 & 0.02 & 0.05 & 0.08 & 1.19 & 1.39 & 34.48 \\ 
 &   3 & 0.00 & 0.01 & 0.07 & 0.08 & 1.72 & 2.00 & 52.14 \\ 
\midrule
 \multirow{3}{5em}{fall} &   1 & 0.16 & 0.22 & 0.14 & 0.32 & 4.53 & 4.39 & 4.32 \\ 
 &   2 & 0.14 & 0.20 & 0.17 & 0.35 & 5.82 & 5.65 & 5.70 \\ 
 &   3 & 0.14 & 0.22 & 0.06 & 0.58 & 6.03 & 5.94 & 6.10 \\
 \midrule
 \multirow{3}{5em}{ksl} &   1 & 0.00 & 0.03 & 0.03 & 10.33 & 10.11 & 226.57 &  \\ 
 &   2 & 0.01 & 0.02 & 0.06 & 11.33 & 10.88 & 442.58 &  \\ 
 &   3 & 0.00 & 0.02 & 0.12 & 12.80 & 12.15 & 751.77 &  \\
\midrule
\multirow{3}{5em}{mathmarks} &   1 & 0.00 & 0.01 & 0.00 & 0.02 & 1.32 & 0.42 & 1.25 \\ 
 &   2 & 0.00 & 0.00 & 0.03 & 0.05 & 4.42 & 1.50 & 4.31 \\ 
 &   3 & 0.00 & 0.02 & 0.49 & 0.44 & 54.95 & 18.20 & 54.46 \\ 
\midrule
 \multirow{3}{5em}{phd} &   1 & 0.00 & 0.02 & 0.02 & 0.23 & 0.92 & 3.53 & 10.58 \\ 
&   2 & 0.01 & 0.02 & 0.04 & 0.47 & 1.78 & 7.00 & 23.11 \\ 
 &   3 & 0.00 & 0.01 & 0.11 & 1.30 & 4.59 & 17.28 & 68.21 \\
 \midrule
 \multirow{3}{5em}{titanic} &   1 & 0.00 & 0.00 & 0.02 & 0.01 & 0.19 & 0.17 & 0.19 \\ 
 &   2 & 0.00 & 0.02 & 0.02 & 0.03 & 0.45 & 0.41 & 0.43 \\ 
 &   3 & 0.00 & 0.01 & 0.05 & 0.06 & 0.95 & 0.97 & 0.93 \\ 
   \bottomrule
\end{tabular}
}
\label{table:time}
\end{table}
Table \ref{table:time} reports the computational times for all models learned. As expected, time increases with $k$, the number of parents, since the model learned is more complex. The DAG algorithm is of course the fastest since it considers the smallest model class. Algorithms for ALDAGs with a fixed order (LV and CMI) are comparably fast, whilst those without a fixed order are the slowest, but still feasible for all datasets. In conclusion, it seems that the ORD1 algorithm (which considers all compatible orders to a BN learned with the tabu procedure) gives a fair compromise between fit and speed of computation. 

Last we report in Table \ref{table:edges} the number of edges having labels total or of some other type. By default, all BN models have all total edge labels. ALDAGs with one parent per vertex only ($k=1$) have often only total edge labels. Models with a better fit, corresponding to ALDAGs with $k=2,3$ learned without a fixed order, have more non-total edge labels, thus highlighting the need to learn more flexible models embedding asymmetric independences. 

\begin{table}
\centering
\caption{Number of edges with label total (left number) and non-total label (right number) for the learned models in Table \ref{table:bic}.} 
\scalebox{0.6}{
\begin{tabular}{lcccccccc}
  \toprule
data & k & DAG & LV & CMI & ORD1 & ORD2 & ORD3 & ALL \\ 
  \midrule
  \multirow{3}{5em}{asia} &   1 & (6, 0) & (6, 0) & (5, 0) & (5, 0) & (5, 0) & (6, 0) &  \\ 
 &   2 & (7, 0) & (3, 4) & (2, 6) & (2, 6) & (2, 6) & (1, 8) &  \\ 
 &   3 & (7, 0) & (3, 4) & (0, 12) & (0, 12) & (0, 12) & (0, 12) &  \\ 
 \midrule
  \multirow{3}{5em}{cachexia} &   1 & (6, 0) & (4, 2) & (3, 3) & (4, 2) & (2, 4) & (4, 2) &  \\ 
 &   2 & (6, 0) & (4, 2) & (1, 10) & (1, 10) & (0, 11) & (0, 11) &  \\ 
 &   3 & (6, 0) & (4, 2) & (1, 14) & (1, 14) & (0, 15) & (0, 15) &  \\ 
 \midrule
  \multirow{3}{5em}{chds} &   1 & (3, 0) & (2, 1) & (2, 1) & (2, 1) & (1, 2) & (1, 1) & (1, 2) \\ 
&   2 & (3, 0) & (2, 1) & (1, 3) & (2, 1) & (1, 2) & (1, 2) & (1, 2) \\ 
 &   3 & (3, 0) & (2, 1) & (1, 5) & (2, 3) & (0, 4) & (0, 4) & (0, 4) \\ 
 \midrule
 \multirow{3}{5em}{coronary} &   1 & (5, 0) & (5, 0) & (5, 0) & (5, 0) & (5, 0) & (4, 0) & (5, 0) \\ 
 &   2 & (8, 0) & (3, 5) & (2, 7) & (2, 7) & (2, 7) & (1, 7) & (1, 8) \\ 
&   3 & (8, 0) & (4, 4) & (2, 10) & (2, 9) & (2, 9) & (1, 10) & (2, 9) \\ 
\midrule
  \multirow{3}{5em}{fall} &   1 & (3, 0) & (2, 1) & (2, 1) & (2, 1) & (2, 1) & (2, 1) & (2, 1) \\ 
&   2 & (5, 0) & (0, 5) & (1, 4) & (0, 5) & (0, 5) & (0, 5) & (0, 5) \\ 
 &   3 & (5, 0) & (0, 5) & (1, 5) & (0, 6) & (0, 6) & (0, 6) & (0, 6) \\ 
 \midrule
 \multirow{3}{5em}{ksl} &   1 & (8, 0) & (8, 0) & (8, 0) & (7, 0) & (7, 0) & (7, 0) &  \\ 
&   2 & (12, 0) & (3, 9) & (2, 12) & (2, 11) & (2, 11) & (1, 13) &  \\ 
 &   3 & (12, 0) & (3, 9) & (2, 15) & (0, 18) & (0, 18) & (0, 18) &  \\ 
 \midrule
   \multirow{3}{5em}{mathmarks} &   1 & (4, 0) & (0, 4) & (0, 4) & (0, 4) & (0, 4) & (1, 3) & (0, 4) \\ 
  &   2 & (4, 0) & (0, 4) & (0, 7) & (0, 7) & (0, 7) & (0, 7) & (0, 7) \\ 
  &   3 & (4, 0) & (0, 4) & (0, 9) & (0, 9) & (0, 9) & (0, 9) & (0, 9) \\ 
  \midrule
  \multirow{3}{5em}{phd} &   1 & (5, 0) & (4, 1) & (3, 2) & (3, 2) & (3, 2) & (4, 1) & (3, 2) \\ 
 &   2 & (6, 0) & (3, 3) & (3, 4) & (3, 6) & (3, 6) & (2, 6) & (3, 6) \\ 
 &   3 & (6, 0) & (3, 3) & (2, 9) & (2, 9) & (2, 9) & (1, 10) & (2, 7) \\ 
 \midrule
  \multirow{3}{5em}{titanic} &   1 & (3, 0) & (1, 2) & (1, 2) & (1, 2) & (1, 2) & (1, 2) & (1, 2) \\ 
  &   2 & (5, 0) & (0, 5) & (0, 5) & (0, 5) & (0, 5) & (0, 5) & (0, 5) \\ 
 &   3 & (5, 0) & (0, 5) & (0, 6) & (0, 6) & (0, 6) & (0, 6) & (0, 6) \\ 
   \bottomrule
\end{tabular}
}
\label{table:edges}
\end{table}

\subsection{Simulation study}

We next perform a simulation study to evaluate the feasibility of the proposed  approach and to demonstrate its superiority to standard BN algorithms under the assumption that the true model is an ALDAG.  We simulate data from randomly generated ALDAGs with different degrees of complexity: number of variables ($p\in \{4,5,6\}$); number of parents per variable ($k \in \{1,2,3\}$); number of stages per variable in the associated staged tree ($t \in \{2,3,4\}$); sample size ($N \in \{250, 500, 1000, 3000, 5000, 10000\}$). 
 For each parameters' combination, we perform $20$ repetitions of the  experiment each time randomly shuffling the order of the variables to  eliminate any possible bias of the search heuristics. The estimation procedures are the same as those of the previous section.

\begin{figure}
    \centering
    \includegraphics{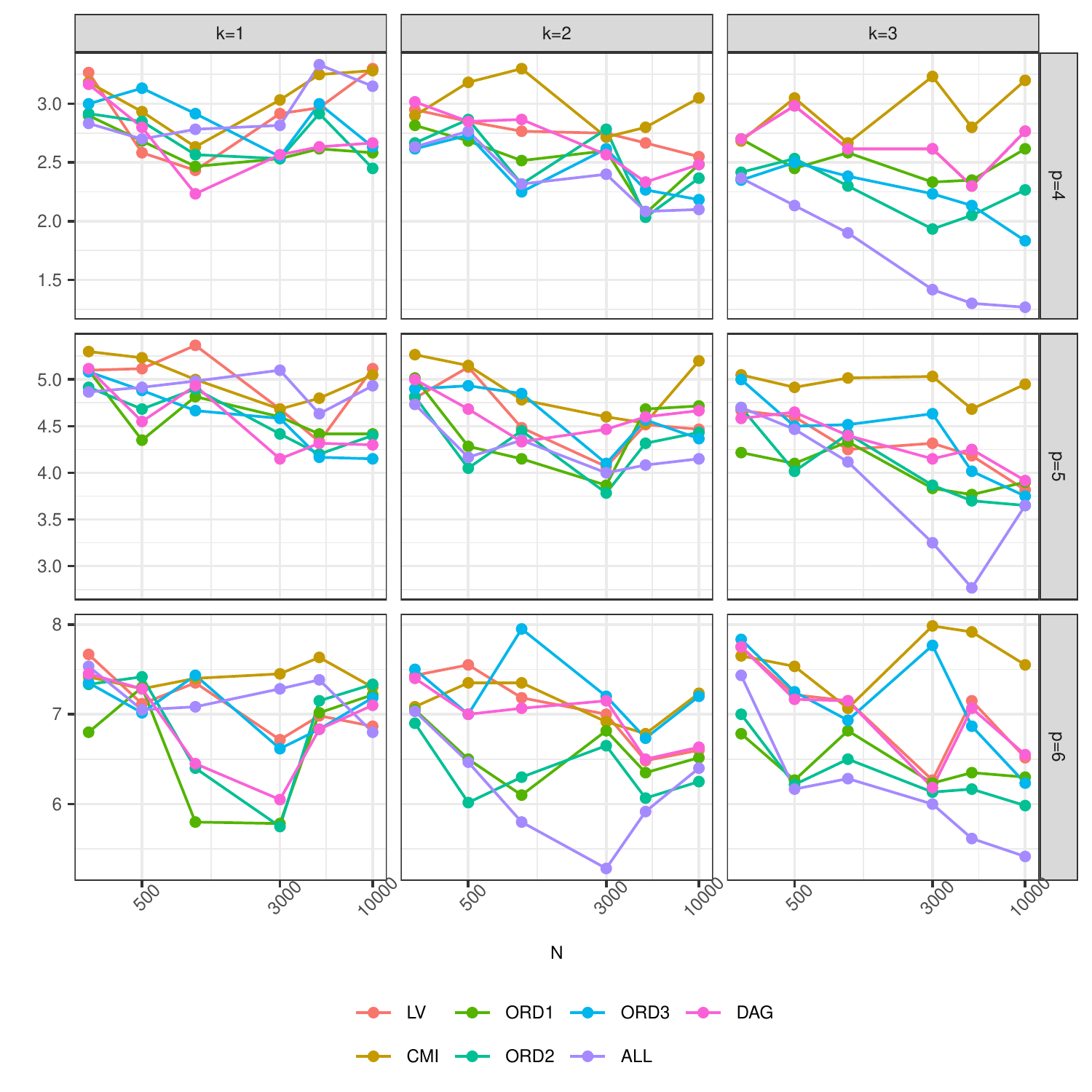}
    \caption{Kendall tau distance between the estimated and true model for d.}
    \label{fig:sim1}
\end{figure}

We first investigate whether the algorithms are capable of retrieving the correct ordering of the variables used for simulating the ALDAG. This is often of interest if the relationship between the variables is assumed to be causal \cite{Pearl2009}. Figure \ref{fig:sim1} reports the Kendall tau distance between the variable orderings as a function of the sample size $N$ for each combination of number of variables $p$ and number of parents $p$. The Kendall tau distance is computed between the true order used to simulate the ALDAG and the estimated order with the implementation 
in the \texttt{PerMallows} R package \cite{permallows}. For the case $k=1$ we can see that there is almost no difference between the approaches in retrieving the correct ordering of the variables. For $k=3$, the algorithm using every possible order outperforms the others as expected, but ORD2 and ORD3 give better results than the standard DAG approach. Not surprisingly, the CMI algorithm, which takes an arbitrary ordering of the variables has the worse performance of all.

\begin{figure}
    \centering
    \includegraphics{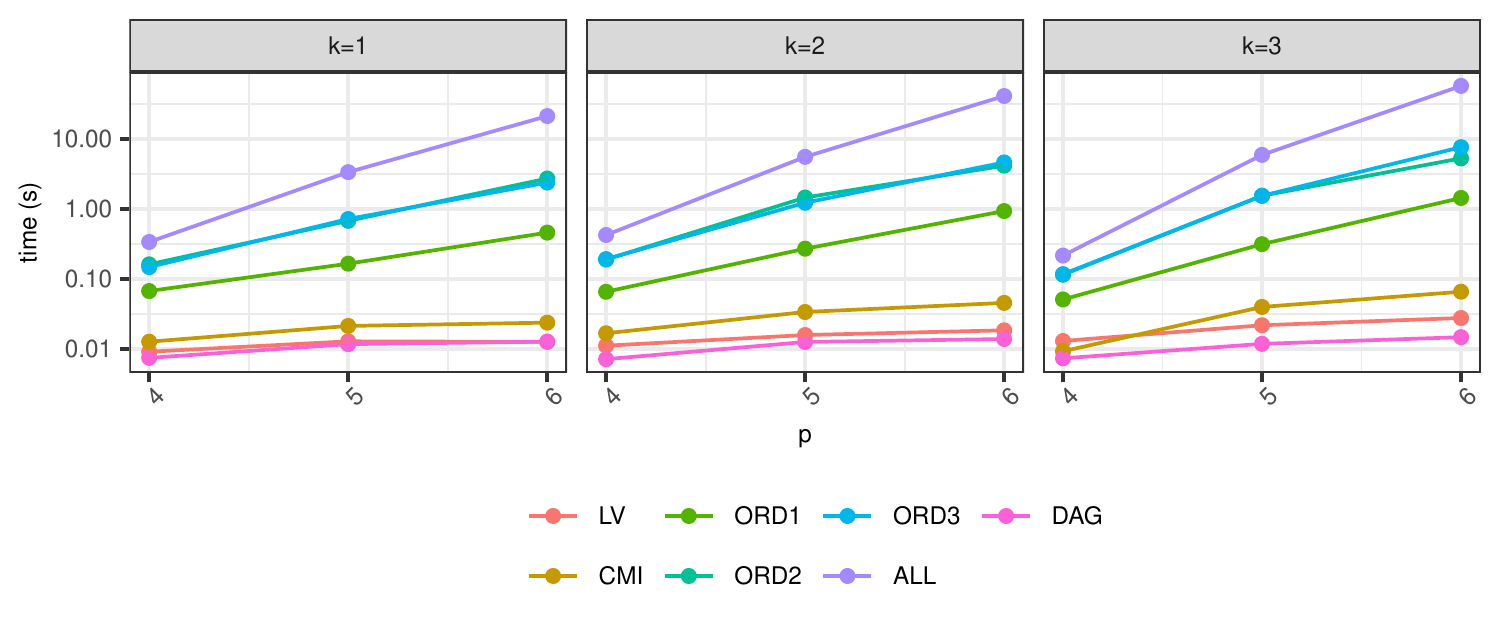}
    \caption{Computational time (in seconds) as a function of the number of variables $p$ for the simulation experiment.}
    \label{fig:exp2}
\end{figure}

Figure \ref{fig:exp2} reports the average time required for the estimation of the models as a function of time. As expected, the DAG approach is the fastest but with a time comparable to that of the fixed-order approach. The algorithm investigating every possible order is the slowest: however, on average it takes only ten seconds in the case of six variables. Notice that the number of parents $k$ does not seem to have a major effect on the computational times.

The average BIC for every combination of number of variables ($p$), number of parents ($k$), and sample size ($N$) was also computed for every estimating procedure (not reported here for space). It showed that, as the sample size increases, ALDAG routines without a fixed order provide a better fit than the standard DAG algorithm. The difference in fit also increases with $p$ and $k$, as seen in the computational study reported above.

\section{Data application: The fear of COVID-19 scale}
\label{sec:data}

The COVID-19 pandemic shook the world and changed the habits of most individuals due to months-long lockdowns of educational and nonessential business activities. An event of this scale and criticality was unprecedented in
the lifespan of basically all citizens worldwide. Many aspects of the COVID-19 pandemic, such as uncertainty over patient outcomes, familiarity with infected people, and mandatory change of habits have led many individuals across the globe to experience a generalized sense of fear \cite{huang}. For previous epidemics, fear has been shown to be associated with other psychological disorders such as anxiety and depression \cite{ford}, and these have also been shown to be strongly related to social isolation \cite{santini}, which in the case of the COVID-19 pandemic has been imposed by social distancing governmental policies.

To quantify and measure the fear of individuals about the COVID-19 pandemic, the Fear of COVID-19 Scale was developed  in \cite{scale}, consisting of the following seven items:
\begin{itemize}
    \item I am most afraid of COVID-19 (Fear).
    \item  It makes me uncomfortable to think about COVID-19 (Think).
  \item My hands become clammy when I think about COVID-19 (Hands).
 \item I am afraid of losing my life because of COVID-19 (Life).
\item When watching news and stories about COVID-19 on social media, I become nervous
or anxious (Media).
\item I cannot sleep because I am worried about getting COVID-19 (Sleep).
 \item My heart races or palpitates when I think about getting COVID-19 (Heart).
\end{itemize}
Participants indicate their level of agreement with the statements using a five-item Likert-type scale. Answers included “strongly disagree,” “disagree,” “neither agree nor disagree,”
“agree,” and “strongly agree”. The minimum score possible for each question is 1, and the
maximum is 5. A total score is calculated by adding up each item score (ranging from 7 to 35).
The higher the score, the greater the fear of COVID-19. Because of the generalized sense of fear associated with the COVID-19 pandemic, the scale has been used and validated in many other countries, including China \cite{china}, Italy \cite{italy} and Spain \cite{spain}. 

Data from the Italian validation of the Fear of COVID-19 Scale is available from \cite{data} and consists of the answers of 149 individuals to the seven items of the scale together with information about their age and gender. Age was dichotomized using the equal-frequency method and the answer to the items of the scale were transformed into ternary variables: disagree (including strongly disagree and disagree), neither (neither agree nor disagree), and agree (including agree and strongly agree).

The analysis aims to understand the effect of the two demographic factors on the answers to the fear scale as well as the relationship between the items of the scale. To this end, DAG and ALDAG models are learned from the data using various methodologies: DAGs using the tabu algorithm; ALDAGs with the routine of \cite{leonelli2022highly}; ALDAGs with a fixed order as in Section \ref{sec:fixed}; ALDAGs considering all compatible orders to an initial BN. The other approaches are not considered here because with a total of nine variables there would be around 362880 orders to consider, and thus computationally unfeasible. Notice however that the experiments of the previous section showcased that the procedure of learning ALDAGs using all compatible orders to an initial BN is competitive in terms of fit to more nuanced approaches. Each of the considered algorithms is evaluated with $k=1,2,3$ number of parents. Furthermore, for all algorithms, only outputs where age and gender have no items' parents are considered, since our interest is in the effect of the demographic variables on the way individuals responded.

\begin{table}
\centering
\caption{Number of edges with label total (left number) and non-total label (right number) for the learned models in Table \ref{table:bic}.} 
\scalebox{0.9}{
\begin{tabular}{ccccc}
  \toprule
k & DAG & LV & CMI & ORD1 \\
\midrule
1 &  3275.158 & 3264.242 &3337.566&3264.242\\
2 & 3275.158 &3264.242 & 3251.970& 3191.428\\
3 & 3275.158 &3264.242 & 3225.318 &\textbf{3166.348}\\
\bottomrule
  \end{tabular}
  }
  \label{table}
  \end{table}
  
The BIC of all learned models are reported in Table \ref{table}. It can be noticed that all DAGs are such that any vertex has at most one parent since all BICs are the same. Consequently, also all ALDAGs learned with the approach of \cite{leonelli2022highly} are such that any vertex has at most one parent. The best scoring model is the one with three parents and the order chosen from those compatible with a starting BN learned with the tabu algorithm.

\begin{figure}
    \centering
    \includegraphics[scale=0.8]{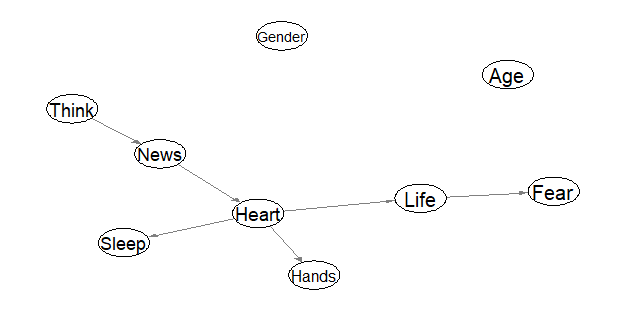}
    \caption{BN learned with the tabu algorithm using the Fear of COVID-19 Scale data.}
    \label{bn}
\end{figure}

Figure \ref{bn} shows the learned BN. It can be seen that indeed every variable has at most one parent. Importantly, it shows that the two demographic variables do not affect how individuals responded to the items on the scale.

\begin{figure}
    \centering
    \includegraphics[scale=0.7]{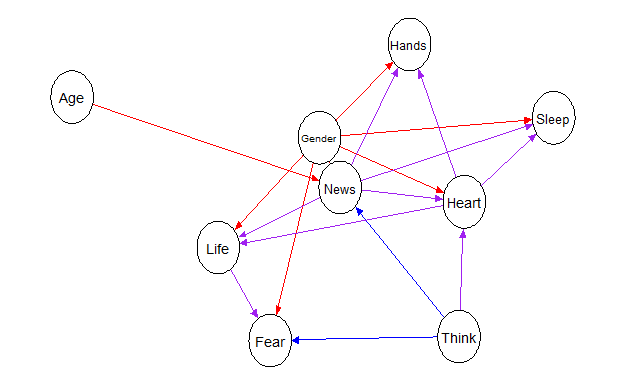}
    \caption{Best scoring ALDAG for the Fear of COVID-19 Scale data. The edge coloring is: red - context; blue - partial; violet - context/partial; green - local; black - total.}
    \label{aldag}
\end{figure}

The best-scoring ALDAG reported in Figure \ref{aldag} reveals that the strict independence assumption between the demographic variables and the items of the scale is not tenable. Age has a direct effect on the item News, whilst Gender has a direct effect on all items except for Think. Furthermore, the model shows a much wider dependence between the items of the scale since it includes all edges of the DAG in Figure \ref{bn} (although with a different label) as well as additional ones. Interestingly, no edges were given a label total, and there are seven edges context, two partial, and eight context/partial.

Symmetric conditional independences can be read from the ALDAG as in standard BNs. For instance, the network implies that, given Gender, Life and Think, Age and the remaining items do not affect how individuals answer the Fear item. In other words, the Fear item, which provides an overview of the level of fear of an individual, is only directly affected by gender, and by how the Life and Think items were answered.

\begin{figure}
\centering
\scalebox{0.6}{
\begin{tikzpicture}
\renewcommand{\xx}{2.5}
\renewcommand{\yy}{0.75}
\node (v0) at (0*\xx,0*\yy) {\sage{white}{0}};
\node (v1) at (1*\xx,-9*\yy) {\sage{white}{1}};
\node (v2) at (1*\xx,9*\yy) {\sage{white}{2}};
\node (v3) at (2*\xx,-15*\yy) {\sage{white}{3}};
\node (v4) at (2*\xx,-9*\yy) {\sage{white}{4}};
\node (v5) at (2*\xx,-3*\yy) {\sage{white}{5}};
\node (v6) at (2*\xx,3*\yy) {\sage{white}{6}};
\node (v7) at (2*\xx,9*\yy) {\sage{white}{7}};
\node (v8) at (2*\xx,15*\yy) {\sage{white}{8}};
\node (v9) at (3*\xx,-16.5*\yy) {\sage{cyan}{9}};
\node (v10) at (3*\xx,-13.5*\yy) {\sage{cyan}{10}};
\node (v11) at (3*\xx,-10.5*\yy) {\sage{green}{11}};
\node (v12) at (3*\xx,-7.5*\yy) {\sage{cyan}{12}};
\node (v13) at (3*\xx,-4.5*\yy) {\sage{orange}{13}};
\node (v14) at (3*\xx,-1.5*\yy) {\sage{orange}{14}};
\node (v15) at (3*\xx,1.5*\yy) {\sage{cyan}{15}};
\node (v16) at (3*\xx,4.5*\yy) {\sage{cyan}{16}};
\node (v17) at (3*\xx,7.5*\yy) {\sage{cyan}{17}};
\node (v18) at (3*\xx,10.5*\yy) {\sage{cyan}{18}};
\node (v19) at (3*\xx,13.5*\yy) {\sage{orange}{19}};
\node (v20) at (3*\xx,16.5*\yy) {\sage{orange}{20}};
\node (v21) at (4*\xx,-17.5*\yy) {\leaf};
\node (v22) at (4*\xx,-16.5*\yy) {\leaf};
\node (v23) at (4*\xx,-15.5*\yy) {\leaf};
\node (v24) at (4*\xx,-14.5*\yy) {\leaf};
\node (v25) at (4*\xx,-13.5*\yy) {\leaf};
\node (v26) at (4*\xx,-12.5*\yy) {\leaf};
\node (v27) at (4*\xx,-11.5*\yy) {\leaf};
\node (v28) at (4*\xx,-10.5*\yy) {\leaf};
\node (v29) at (4*\xx,-9.5*\yy) {\leaf};
\node (v30) at (4*\xx,-8.5*\yy) {\leaf};
\node (v31) at (4*\xx,-7.5*\yy) {\leaf};
\node (v32) at (4*\xx,-6.5*\yy) {\leaf};
\node (v33) at (4*\xx,-5.5*\yy) {\leaf};
\node (v34) at (4*\xx,-4.5*\yy) {\leaf};
\node (v35) at (4*\xx,-3.5*\yy) {\leaf};
\node (v36) at (4*\xx,-2.5*\yy) {\leaf};
\node (v37) at (4*\xx,-1.5*\yy) {\leaf};
\node (v38) at (4*\xx,-0.5*\yy) {\leaf};
\node (v39) at (4*\xx,0.5*\yy) {\leaf};
\node (v40) at (4*\xx,1.5*\yy) {\leaf};
\node (v41) at (4*\xx,2.5*\yy) {\leaf};
\node (v42) at (4*\xx,3.5*\yy) {\leaf};
\node (v43) at (4*\xx,4.5*\yy) {\leaf};
\node (v44) at (4*\xx,5.5*\yy) {\leaf};
\node (v45) at (4*\xx,6.5*\yy) {\leaf};
\node (v46) at (4*\xx,7.5*\yy) {\leaf};
\node (v47) at (4*\xx,8.5*\yy) {\leaf};
\node (v48) at (4*\xx,9.5*\yy) {\leaf};
\node (v49) at (4*\xx,10.5*\yy) {\leaf};
\node (v50) at (4*\xx,11.5*\yy) {\leaf};
\node (v51) at (4*\xx,12.5*\yy) {\leaf};
\node (v52) at (4*\xx,13.5*\yy) {\leaf};
\node (v53) at (4*\xx,14.5*\yy) {\leaf};
\node (v54) at (4*\xx,15.5*\yy) {\leaf};
\node (v55) at (4*\xx,16.5*\yy) {\leaf};
\node (v56) at (4*\xx,17.5*\yy) {\leaf};
\draw[->] (v0) -- node [below, sloped] {\tiny{young}} (v1);
\draw[->] (v0) -- node [below, sloped] {\tiny{adult}} (v2);
\draw[->] (v1) --  node [below, sloped] {\tiny{disagree}} (v3);
\draw[->] (v1) --  node [above, sloped] {\tiny{neither}} (v4);
\draw[->] (v1) --  node [above, sloped] {\tiny{agree}} (v5);
\draw[->] (v2) --  node [below, sloped] {\tiny{disagree}} (v6);
\draw[->] (v2) --  node [above, sloped] {\tiny{neither}} (v7);
\draw[->] (v2) --  node [above, sloped] {\tiny{agree}} (v8);
\draw[->] (v3) --  node [below, sloped] {\tiny{female}} (v9);
\draw[->] (v3) --  node [above, sloped] {\tiny{male}} (v10);
\draw[->] (v4) --  node [below, sloped] {\tiny{female}} (v11);
\draw[->] (v4) --  node [above, sloped] {\tiny{male}} (v12);
\draw[->] (v5) --  node [below, sloped] {\tiny{female}} (v13);
\draw[->] (v5) --  node [above, sloped] {\tiny{male}} (v14);
\draw[->] (v6) --  node [below, sloped] {\tiny{female}} (v15);
\draw[->] (v6) --  node [above, sloped] {\tiny{male}} (v16);
\draw[->] (v7) --  node [below, sloped] {\tiny{female}} (v17);
\draw[->] (v7) --  node [above, sloped] {\tiny{male}} (v18);
\draw[->] (v8) --  node [below, sloped] {\tiny{female}} (v19);
\draw[->] (v8) --  node [above, sloped] {\tiny{male}} (v20);
\draw[->] (v9) --  node [below, sloped] {\tiny{disagree}} (v21);
\draw[->] (v9) --  node [above, sloped] {\tiny{neither}} (v22);
\draw[->] (v9) --  node [above, sloped] {\tiny{agree}} (v23);
\draw[->] (v10) --  node [below, sloped] {\tiny{disagree}} (v24);
\draw[->] (v10) --  node [above, sloped] {\tiny{neither}} (v25);
\draw[->] (v10) --  node [above, sloped] {\tiny{agree}} (v26);
\draw[->] (v11) --  node [below, sloped] {\tiny{disagree}} (v27);
\draw[->] (v11) --  node [above, sloped] {\tiny{neither}} (v28);
\draw[->] (v11) --  node [above, sloped] {\tiny{agree}} (v29);
\draw[->] (v12) --  node [below, sloped] {\tiny{disagree}} (v30);
\draw[->] (v12) --  node [above, sloped] {\tiny{neither}} (v31);
\draw[->] (v12) --  node [above, sloped] {\tiny{agree}} (v32);
\draw[->] (v13) --  node [below, sloped] {\tiny{disagree}} (v33);
\draw[->] (v13) --  node [above, sloped] {\tiny{neither}} (v34);
\draw[->] (v13) --  node [above, sloped] {\tiny{agree}} (v35);
\draw[->] (v14) --  node [below, sloped] {\tiny{disagree}} (v36);
\draw[->] (v14) --  node [above, sloped] {\tiny{neither}} (v37);
\draw[->] (v14) --  node [above, sloped] {\tiny{agree}} (v38);
\draw[->] (v15) --  node [below, sloped] {\tiny{disagree}} (v39);
\draw[->] (v15) --  node [above, sloped] {\tiny{neither}} (v40);
\draw[->] (v15) --  node [above, sloped] {\tiny{agree}} (v41);
\draw[->] (v16) --  node [below, sloped] {\tiny{disagree}} (v42);
\draw[->] (v16) --  node [above, sloped] {\tiny{neither}} (v43);
\draw[->] (v16) --  node [above, sloped] {\tiny{agree}} (v44);
\draw[->] (v17) --  node [below, sloped] {\tiny{disagree}} (v45);
\draw[->] (v17) --  node [above, sloped] {\tiny{neither}} (v46);
\draw[->] (v17) --  node [above, sloped] {\tiny{agree}} (v47);
\draw[->] (v18) --  node [below, sloped] {\tiny{disagree}} (v48);
\draw[->] (v18) --  node [above, sloped] {\tiny{neither}} (v49);
\draw[->] (v18) --  node [above, sloped] {\tiny{agree}} (v50);
\draw[->] (v19) --  node [below, sloped] {\tiny{disagree}} (v51);
\draw[->] (v19) --  node [above, sloped] {\tiny{neither}} (v52);
\draw[->] (v19) --  node [above, sloped] {\tiny{agree}} (v53);
\draw[->] (v20) --  node [below, sloped] {\tiny{disagree}} (v54);
\draw[->] (v20) --  node [above, sloped] {\tiny{neither}} (v55);
\draw[->] (v20) --  node [above, sloped] {\tiny{agree}} (v56);
\end{tikzpicture}
}

\caption{Dependence subtree for the variable News with parents (Age, Think, Gender). \label{dep1}}
\end{figure}
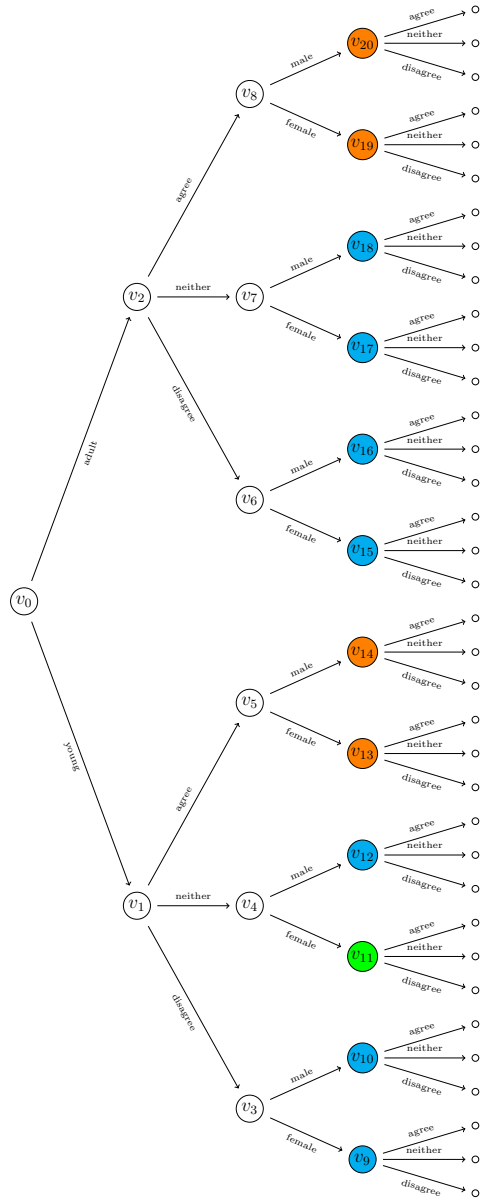

The labeling of the edges provides further information about the relationship between a variable and its parents. In \cite{Varando2022aldag}, the so-called \emph{dependence subtree} was introduced to retrieve  the actual dependence structure between a variable and its parents from the ALDAG. This is shown in Figure \ref{dep1} for the variable News, which is directly affected by the demographic variables as well as the Think item. The coloring of the vertices associated with the News variable ($v_9-v_{20})$ represents asymmetric conditional independence as for staged trees. The coloring shows that for individuals who answered agree or disagree to the Think item, gender and age do not affect the way they answered News. It also holds that gender is independent of News for the adult group who answered Think = neither.

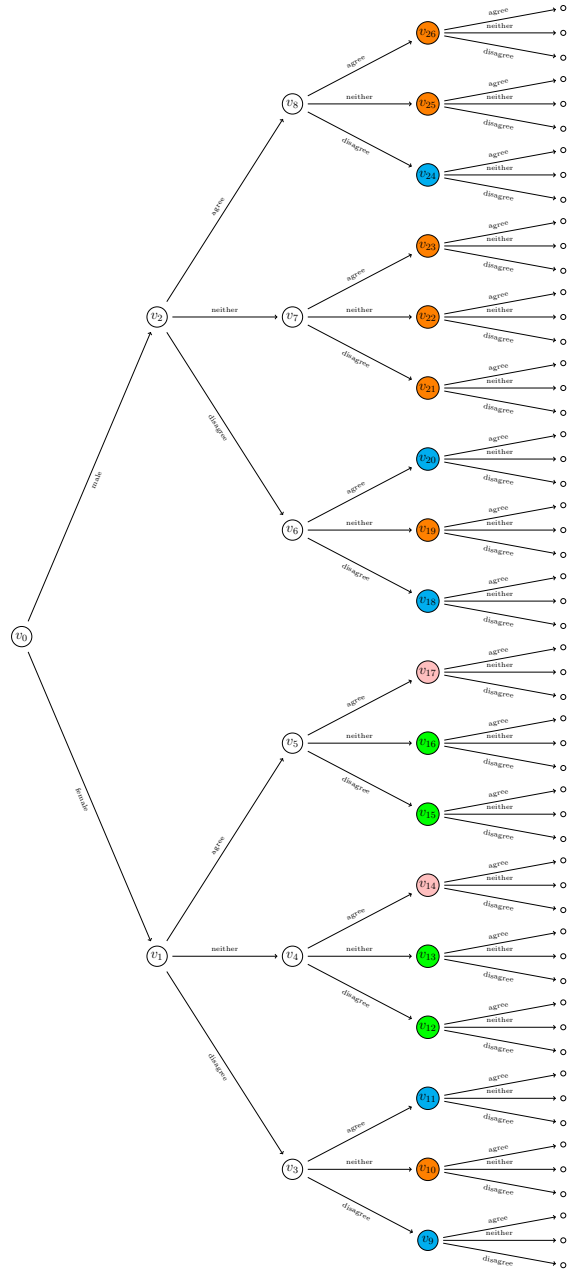
\begin{figure}
\centering
\scalebox{0.45}{
\begin{tikzpicture}
\renewcommand{\xx}{4}
\renewcommand{\yy}{1.05}
\node (v0) at (0*\xx,0*\yy) {\sage{white}{0}};
\node (v1) at (1*\xx,-9*\yy) {\sage{white}{1}};
\node (v2) at (1*\xx,9*\yy) {\sage{white}{2}};
\node (v3) at (2*\xx,-15*\yy) {\sage{white}{3}};
\node (v4) at (2*\xx,-9*\yy) {\sage{white}{4}};
\node (v5) at (2*\xx,-3*\yy) {\sage{white}{5}};
\node (v6) at (2*\xx,3*\yy) {\sage{white}{6}};
\node (v7) at (2*\xx,9*\yy) {\sage{white}{7}};
\node (v8) at (2*\xx,15*\yy) {\sage{white}{8}};
\node (v9) at (3*\xx,-17*\yy) {\sage{cyan}{9}};
\node (v10) at (3*\xx,-15*\yy) {\sage{orange}{10}};
\node (v11) at (3*\xx,-13*\yy) {\sage{cyan}{11}};
\node (v12) at (3*\xx,-11*\yy) {\sage{green}{12}};
\node (v13) at (3*\xx,-9*\yy) {\sage{green}{13}};
\node (v14) at (3*\xx,-7*\yy) {\sage{pink}{14}};
\node (v15) at (3*\xx,-5*\yy) {\sage{green}{15}};
\node (v16) at (3*\xx,-3*\yy) {\sage{green}{16}};
\node (v17) at (3*\xx,-1*\yy) {\sage{pink}{17}};
\node (v18) at (3*\xx,1*\yy) {\sage{cyan}{18}};
\node (v19) at (3*\xx,3*\yy) {\sage{orange}{19}};
\node (v20) at (3*\xx,5*\yy) {\sage{cyan}{20}};
\node (v21) at (3*\xx,7*\yy) {\sage{orange}{21}};
\node (v22) at (3*\xx,9*\yy) {\sage{orange}{22}};
\node (v23) at (3*\xx,11*\yy) {\sage{orange}{23}};
\node (v24) at (3*\xx,13*\yy) {\sage{cyan}{24}};
\node (v25) at (3*\xx,15*\yy) {\sage{orange}{25}};
\node (v26) at (3*\xx,17*\yy) {\sage{orange}{26}};
\node (v27) at (4*\xx,-17.7*\yy) {\leaf};
\node (v28) at (4*\xx,-17*\yy) {\leaf};
\node (v29) at (4*\xx,-16.3*\yy) {\leaf};
\node (v30) at (4*\xx,-15.7*\yy) {\leaf};
\node (v31) at (4*\xx,-15*\yy) {\leaf};
\node (v32) at (4*\xx,-14.3*\yy) {\leaf};
\node (v33) at (4*\xx,-13.7*\yy) {\leaf};
\node (v34) at (4*\xx,-13*\yy) {\leaf};
\node (v35) at (4*\xx,-12.3*\yy) {\leaf};
\node (v36) at (4*\xx,-11.7*\yy) {\leaf};
\node (v37) at (4*\xx,-11*\yy) {\leaf};
\node (v38) at (4*\xx,-10.3*\yy) {\leaf};
\node (v39) at (4*\xx,-9.7*\yy) {\leaf};
\node (v40) at (4*\xx,-9*\yy) {\leaf};
\node (v41) at (4*\xx,-8.3*\yy) {\leaf};
\node (v42) at (4*\xx,-7.7*\yy) {\leaf};
\node (v43) at (4*\xx,-7*\yy) {\leaf};
\node (v44) at (4*\xx,-6.3*\yy) {\leaf};
\node (v45) at (4*\xx,-5.7*\yy) {\leaf};
\node (v46) at (4*\xx,-5*\yy) {\leaf};
\node (v47) at (4*\xx,-4.3*\yy) {\leaf};
\node (v48) at (4*\xx,-3.7*\yy) {\leaf};
\node (v49) at (4*\xx,-3*\yy) {\leaf};
\node (v50) at (4*\xx,-2.3*\yy) {\leaf};
\node (v51) at (4*\xx,-1.7*\yy) {\leaf};
\node (v52) at (4*\xx,-1*\yy) {\leaf};
\node (v53) at (4*\xx,-0.3*\yy) {\leaf};
\node (v54) at (4*\xx,0.3*\yy) {\leaf};
\node (v55) at (4*\xx,1*\yy) {\leaf};
\node (v56) at (4*\xx,1.7*\yy) {\leaf};
\node (v57) at (4*\xx,2.3*\yy) {\leaf};
\node (v58) at (4*\xx,3*\yy) {\leaf};
\node (v59) at (4*\xx,3.7*\yy) {\leaf};
\node (v60) at (4*\xx,4.3*\yy) {\leaf};
\node (v61) at (4*\xx,5*\yy) {\leaf};
\node (v62) at (4*\xx,5.7*\yy) {\leaf};
\node (v63) at (4*\xx,6.3*\yy) {\leaf};
\node (v64) at (4*\xx,7*\yy) {\leaf};
\node (v65) at (4*\xx,7.7*\yy) {\leaf};
\node (v66) at (4*\xx,8.3*\yy) {\leaf};
\node (v67) at (4*\xx,9*\yy) {\leaf};
\node (v68) at (4*\xx,9.7*\yy) {\leaf};
\node (v69) at (4*\xx,10.3*\yy) {\leaf};
\node (v70) at (4*\xx,11*\yy) {\leaf};
\node (v71) at (4*\xx,11.7*\yy) {\leaf};
\node (v72) at (4*\xx,12.3*\yy) {\leaf};
\node (v73) at (4*\xx,13*\yy) {\leaf};
\node (v74) at (4*\xx,13.7*\yy) {\leaf};
\node (v75) at (4*\xx,14.3*\yy) {\leaf};
\node (v76) at (4*\xx,15*\yy) {\leaf};
\node (v77) at (4*\xx,15.7*\yy) {\leaf};
\node (v78) at (4*\xx,16.3*\yy) {\leaf};
\node (v79) at (4*\xx,17*\yy) {\leaf};
\node (v80) at (4*\xx,17.7*\yy) {\leaf};
\draw[->] (v0) -- node [below, sloped] {\tiny{female}} (v1);
\draw[->] (v0) -- node [below, sloped] {\tiny{male}} (v2);
\draw[->] (v1) --  node [below, sloped] {\tiny{disagree}} (v3);
\draw[->] (v1) --  node [above, sloped] {\tiny{neither}} (v4);
\draw[->] (v1) --  node [above, sloped] {\tiny{agree}} (v5);
\draw[->] (v2) --  node [below, sloped] {\tiny{disagree}} (v6);
\draw[->] (v2) --  node [above, sloped] {\tiny{neither}} (v7);
\draw[->] (v2) --  node [above, sloped] {\tiny{agree}} (v8);
\draw[->] (v3) --  node [below, sloped] {\tiny{disagree}} (v9);
\draw[->] (v3) --  node [above, sloped] {\tiny{neither}} (v10);
\draw[->] (v3) --  node [above, sloped] {\tiny{agree}} (v11);
\draw[->] (v4) --  node [below, sloped] {\tiny{disagree}} (v12);
\draw[->] (v4) --  node [above, sloped] {\tiny{neither}} (v13);
\draw[->] (v4) --  node [above, sloped] {\tiny{agree}} (v14);
\draw[->] (v5) --  node [below, sloped] {\tiny{disagree}} (v15);
\draw[->] (v5) --  node [above, sloped] {\tiny{neither}} (v16);
\draw[->] (v5) --  node [above, sloped] {\tiny{agree}} (v17);
\draw[->] (v6) --  node [below, sloped] {\tiny{disagree}} (v18);
\draw[->] (v6) --  node [above, sloped] {\tiny{neither}} (v19);
\draw[->] (v6) --  node [above, sloped] {\tiny{agree}} (v20);
\draw[->] (v7) --  node [below, sloped] {\tiny{disagree}} (v21);
\draw[->] (v7) --  node [above, sloped] {\tiny{neither}} (v22);
\draw[->] (v7) --  node [above, sloped] {\tiny{agree}} (v23);
\draw[->] (v8) --  node [below, sloped] {\tiny{disagree}} (v24);
\draw[->] (v8) --  node [above, sloped] {\tiny{neither}} (v25);
\draw[->] (v8) --  node [above, sloped] {\tiny{agree}} (v26);
\draw[->] (v9) --  node [below, sloped] {\tiny{disagree}} (v27);
\draw[->] (v9) --  node [above, sloped] {\tiny{neither}} (v28);
\draw[->] (v9) --  node [above, sloped] {\tiny{agree}} (v29);
\draw[->] (v10) --  node [below, sloped] {\tiny{disagree}} (v30);
\draw[->] (v10) --  node [above, sloped] {\tiny{neither}} (v31);
\draw[->] (v10) --  node [above, sloped] {\tiny{agree}} (v32);
\draw[->] (v11) --  node [below, sloped] {\tiny{disagree}} (v33);
\draw[->] (v11) --  node [above, sloped] {\tiny{neither}} (v34);
\draw[->] (v11) --  node [above, sloped] {\tiny{agree}} (v35);
\draw[->] (v12) --  node [below, sloped] {\tiny{disagree}} (v36);
\draw[->] (v12) --  node [above, sloped] {\tiny{neither}} (v37);
\draw[->] (v12) --  node [above, sloped] {\tiny{agree}} (v38);
\draw[->] (v13) --  node [below, sloped] {\tiny{disagree}} (v39);
\draw[->] (v13) --  node [above, sloped] {\tiny{neither}} (v40);
\draw[->] (v13) --  node [above, sloped] {\tiny{agree}} (v41);
\draw[->] (v14) --  node [below, sloped] {\tiny{disagree}} (v42);
\draw[->] (v14) --  node [above, sloped] {\tiny{neither}} (v43);
\draw[->] (v14) --  node [above, sloped] {\tiny{agree}} (v44);
\draw[->] (v15) --  node [below, sloped] {\tiny{disagree}} (v45);
\draw[->] (v15) --  node [above, sloped] {\tiny{neither}} (v46);
\draw[->] (v15) --  node [above, sloped] {\tiny{agree}} (v47);
\draw[->] (v16) --  node [below, sloped] {\tiny{disagree}} (v48);
\draw[->] (v16) --  node [above, sloped] {\tiny{neither}} (v49);
\draw[->] (v16) --  node [above, sloped] {\tiny{agree}} (v50);
\draw[->] (v17) --  node [below, sloped] {\tiny{disagree}} (v51);
\draw[->] (v17) --  node [above, sloped] {\tiny{neither}} (v52);
\draw[->] (v17) --  node [above, sloped] {\tiny{agree}} (v53);
\draw[->] (v18) --  node [below, sloped] {\tiny{disagree}} (v54);
\draw[->] (v18) --  node [above, sloped] {\tiny{neither}} (v55);
\draw[->] (v18) --  node [above, sloped] {\tiny{agree}} (v56);
\draw[->] (v19) --  node [below, sloped] {\tiny{disagree}} (v57);
\draw[->] (v19) --  node [above, sloped] {\tiny{neither}} (v58);
\draw[->] (v19) --  node [above, sloped] {\tiny{agree}} (v59);
\draw[->] (v20) --  node [below, sloped] {\tiny{disagree}} (v60);
\draw[->] (v20) --  node [above, sloped] {\tiny{neither}} (v61);
\draw[->] (v20) --  node [above, sloped] {\tiny{agree}} (v62);
\draw[->] (v21) --  node [below, sloped] {\tiny{disagree}} (v63);
\draw[->] (v21) --  node [above, sloped] {\tiny{neither}} (v64);
\draw[->] (v21) --  node [above, sloped] {\tiny{agree}} (v65);
\draw[->] (v22) --  node [below, sloped] {\tiny{disagree}} (v66);
\draw[->] (v22) --  node [above, sloped] {\tiny{neither}} (v67);
\draw[->] (v22) --  node [above, sloped] {\tiny{agree}} (v68);
\draw[->] (v23) --  node [below, sloped] {\tiny{disagree}} (v69);
\draw[->] (v23) --  node [above, sloped] {\tiny{neither}} (v70);
\draw[->] (v23) --  node [above, sloped] {\tiny{agree}} (v71);
\draw[->] (v24) --  node [below, sloped] {\tiny{disagree}} (v72);
\draw[->] (v24) --  node [above, sloped] {\tiny{neither}} (v73);
\draw[->] (v24) --  node [above, sloped] {\tiny{agree}} (v74);
\draw[->] (v25) --  node [below, sloped] {\tiny{disagree}} (v75);
\draw[->] (v25) --  node [above, sloped] {\tiny{neither}} (v76);
\draw[->] (v25) --  node [above, sloped] {\tiny{agree}} (v77);
\draw[->] (v26) --  node [below, sloped] {\tiny{disagree}} (v78);
\draw[->] (v26) --  node [above, sloped] {\tiny{neither}} (v79);
\draw[->] (v26) --  node [above, sloped] {\tiny{agree}} (v80);
\end{tikzpicture}
}

\caption{Dependence subtree for the variable Life with parents (Gender, News, Hearth). \label{dep2}}
\end{figure}

Next, we analyze the Life item, since it is perhaps the most critical out of the seven items. From standard conditional independence for DAGs, we can conclude that given gender, News, and Heart, Life is independent of Age, Think, Hands and Sleep. The dependence subtree in Figure \ref{dep2} sheds light on the dependence structure between Life and its parents. Compared to the one in Figure \ref{dep1} we can see a slightly less regular structure embedding various asymmetric independences. For instance for males who answered News = neither, Hearth and Life are independent. Similarly, Life is independent of gender given Hearth and News = disagree. 

Such complex patterns of dependence cannot be modeled using BNs, but they provide valuable information about the dependence structure of the data. The goodness of the ALDAG is evidenced by the BIC of the model, which is much lower than that of the BN model.

\section{Discussion}

ALDAGs are an expressive extension of standard DAGs which carry information about additional equalities in the conditional distributions of the model. In this paper, we have introduced efficient learning algorithms for ALDAGs which return interpretable, sparse models, which can be embellished by small-dimensional staged trees. The computational study and the real-world data applications showcase the power of ALDAGs and our learning algorithms in untangling complex, asymmetric dependence structures.

A promising line of research is in extending ALDAGs to carry out casual information about the variables' relationships. Causal discovery using ALDAGs would provide even more detailed information about causal effects than in the case of DAGs. We have started to investigate this topic in \cite{leonelli2022causal}, but a full characterization of causal ALDAGs is still missing.

A different avenue to impose sparsity in the learned model would be to take a Bayesian learning approach. Then the prior distribution over the space of all possible ALDAGs could penalize more complex DAG structures in favor of simpler ones. We are currently investigating the implementation of such an approach and the use of different prior distributions.

\end{document}